\documentclass[11pt]{article}
\usepackage{amsmath}

\usepackage[preprint]{acl}

\usepackage{times}
\usepackage{latexsym}
\usepackage{xspace}

\usepackage[T1]{fontenc}

\usepackage[utf8]{inputenc}

\usepackage{microtype}

\usepackage{inconsolata}

\usepackage{graphicx}

\usepackage{float} 
\usepackage{tabularx} 
\usepackage[table]{xcolor}
\usepackage{multirow}

\usepackage{tcolorbox}
\tcbuselibrary{skins,breakable}

\usepackage{pifont}
\newcommand{\cmark}{\ding{51}} 
\newcommand{\xmark}{\ding{55}} 

\usepackage{float}
\usepackage{tcolorbox} 
\usepackage{listings}

\newlength{\onecolwid}
\AtBeginDocument{%
  \setlength{\onecolwid}{\dimexpr.5\textwidth-\columnsep/2\relax}%
}

\usepackage{xcolor}

\usepackage{algorithm}
\usepackage{algorithmic}

\lstdefinestyle{pythonstyle}{
  language=Python,
  basicstyle=\ttfamily\small,
  keywordstyle=\color{blue},
  stringstyle=\color{orange},
  commentstyle=\color{gray},
  showstringspaces=false,
  frame=single,
  rulecolor=\color{gray!40},
  breaklines=true,
  tabsize=4,
  captionpos=b
}

\newcommand{\method}{\mbox{\textsc{LWE}}\xspace}
\newcommand{\methodselective}{\mbox{\textit{Selective} \textsc{LWE}}\xspace}

\makeatletter
\DeclareRobustCommand\onedot{\futurelet\@let@token\@onedot}
\def\@onedot{\ifx\@let@token.\else.\null\fi\xspace}

\def\eg{\emph{e.g}\onedot} 
\def\ie{\emph{i.e}\onedot} 
 
 \def\vs{\emph{vs}\onedot}

\makeatother

\title{Becoming Experienced Judges: \\
Selective Test-Time Learning for Evaluators}

\author{
Seungyeon Jwa$^{1}$\hspace{0.5em}
Daechul Ahn$^{1}$\hspace{0.5em}
Reokyoung Kim$^{1}$\hspace{0.5em}
Dongyeop Kang$^{2}$\hspace{0.5em}
Jonghyun Choi$^{1}$\vspace{0.5em}\\
{\hspace{0.5em}$^1$Seoul National University\hspace{0.5em}$^2$University of Minnesota} \vspace{0.5em}\\
{\tt\small \{amyj97,daechulahn,reokyoungkim,jonghyunchoi\}@snu.ac.kr dongyeop@umn.edu}
}

\begin{document}
\maketitle

    \begin{abstract}
    Automatic evaluation with large language models, commonly known as \emph{LLM-as-a-judge}, is now standard across reasoning and alignment tasks.  
    Despite evaluating many samples in deployment, these evaluators typically (i) treat each case independently, missing the opportunity to accumulate experience,
    and (ii) rely on a single fixed prompt for all cases, neglecting the need for sample-specific evaluation criteria.
    We introduce \textbf{Learning While Evaluating} (LWE), a framework that allows evaluators to improve \textit{sequentially} at inference time without requiring training or validation sets.  
    LWE maintains an evolving \emph{meta-prompt} that (i) produces sample-specific evaluation instructions and (ii) refines itself through self-generated feedback.
    Furthermore, we propose \methodselective, which updates the meta-prompt only on self-inconsistent cases, focusing computation where it matters most.  
    This selective approach retains the benefits of sequential learning while being far more cost-effective.  
    Across two pairwise comparison benchmarks, \methodselective outperforms strong baselines, empirically demonstrating that evaluators can improve during sequential testing with a simple selective update—learning most from the cases they struggle with.
    \end{abstract}

    \begin{figure}[t]
        \centering
        \includegraphics[width=1\linewidth]{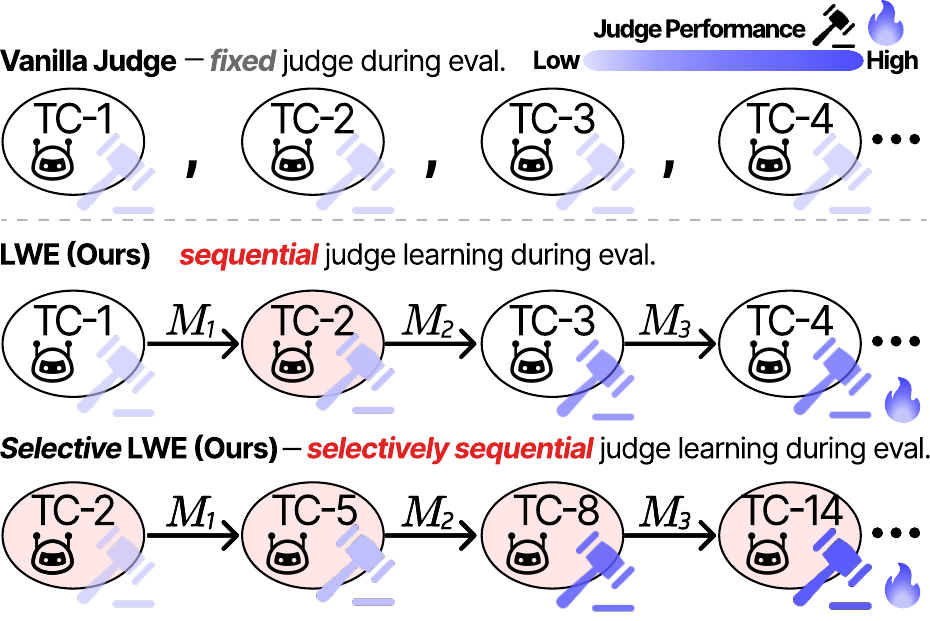}
        \caption{
            \textbf{Comparison of three evaluation approaches.}
            A vanilla judge evaluates each test case (TC) independently using a fixed prompt throughout evaluation (eval.).
            LWE (ours) employs a \emph{meta-prompt} (\textit{M}) that evolves sequentially as the evaluator progresses through test cases, enabling sample-specific tailoring and continual improvement during evaluation.
            \methodselective (ours) further enhances efficiency by updating the meta-prompt only on challenging cases (\eg{}, the red-highlighted TCs 2, 5, 8, 14), preserving performance gains while substantially reducing computational overhead.
            The color gradient illustrates progressive improvement of the judge’s performance over time.
        }
        \vspace{-0.5em}
        \label{fig:lwe_figure1_teaser}
    \end{figure}

    \section{Introduction}
    Large language models (LLMs) and vision-language models (VLMs) are increasingly used as automatic evaluators, commonly referred to as \textit{(V)LLM-as-a-judge}~\cite{zheng2023judging}, in both training and inference stages.
    They guide training as reward models aligning model behavior with human preferences or support iterative model-improvement~\cite{bai2022training, ouyang2022training}, and continue to operate at inference time as automated judges that assess data quality~\cite{grattafiori2024llama}, steer model behavior in inference-time scaling or agentic workflows~\cite{shinn2023reflexion}.

    Despite their growing importance, current evaluators remain surprisingly rigid: they treat test samples independently and typically rely on a single fixed prompt, as illustrated in Figure~\ref{fig:lwe_figure1_teaser}.
    This rigidity prevents two desirable capabilities: tailoring the evaluation criteria to each test case, and improving the evaluator’s judgment quality through use after deployment.
    This motivates our research question: \textit{can evaluators adapt and improve during testing?}  
    Achieving this would provide a highly practical deployment-time capability—evaluators that become better purely through use, without any additional training or external supervision.

    Humans routinely exhibit such test-time improvement.
    A student taking an exam refines their strategy as they progress, and a judge becomes more reliable as they accumulate experience across diverse cases~\cite{flavell1979metacognition, sternberg1985beyond}.
    Recent test-time scaling approaches move in a similar direction by allocating more inference computation per sample, but the extended reasoning they produce is discarded immediately afterward, leaving the thought process inherently sample-independent.
    Thus, current automatic evaluators cannot retain or refine their reasoning \textit{across} cases, raising a methodological question: how can we enable evaluators to learn from their evaluation experience?
        
    To support such learning, an evaluator must be able to retain and use experience from one case to the next.
    Recent progress in non-evaluation settings~\cite{wang2024agent, liu2025contextual, chen2025log} demonstrates that models can improve through sequential inference based on gained information across test cases.
    In particular, \citet{suzgun2025dynamic} introduce a memory prompt that reflects on similar past and current examples to generate the adaptive memory.
    
    These approaches highlight the value of leveraging self-generated insights accumulated during sequential problem solving.  
    However, evaluation requires a different form of knowledge: an evaluator should maintain stable, general evaluation principles while also adapting its judging criteria to each sample.
    To address this, we introduce a \emph{meta-prompt} that explicitly separates general evaluation principles from sample-specific prompts that adapt these principles to each case.
    
    We propose \textbf{Learning While Evaluating} (\method), a framework that enables evaluators to maintain and continually refine their \textit{evaluation insights} as they process test cases.
    At the core of~\method, a meta-prompt serves as a persistent repository of evaluation insights discovered during testing and generates customized evaluation criteria and steps for each new case (Sec.~\ref{sec:lwe}).
    Through self-feedback mechanisms, \method continually improves and expands its meta-prompt during testing, enabling evaluators to learn—rather than merely repeat—their evaluation behavior (Fig.~\ref{fig:method}).
    
    Furthermore, while \method enables evaluators to learn from experience, updating every sample is computationally expensive and often unnecessary, since some test cases are already straightforward and can be accurately judged without additional reasoning.
    To focus updates on cases where more refined evaluation criteria are actually needed, we exploit a test-time–accessible signal that indicates when the evaluator’s fixed judging criteria are insufficient.
    Prior work~\cite{zheng2023judging, shi2024judging, koo2024benchmarking} reports that positional bias can lead a model to give different judgments depending on which answer is presented first, revealing uncertainty or conflict in its internal criteria.
    Leveraging this inconsistency as a test-time signal, we propose \methodselective, which updates the meta-prompt only on such inconsistent cases, retaining the benefits of sequential learning while substantially reducing computational overhead (Sec.~\ref{sec:select_lwe}).
    
    Experiments on two pairwise comparison benchmarks~\cite{li2025vlrewardbench,yasunaga2025multimodalrewardbenchholisticevaluation} show that our framework improves accuracy and consistency over strong baselines while reducing inference cost.
    Overall, we demonstrate that evaluators need not remain static: they can also \textit{learn from experience}, especially from the cases that challenge them the most.

    \begin{figure*}[t]
        \centering
            \includegraphics[width=1\textwidth]{fig_lwe_figure2.pdf}
        \caption{\textbf{Overview of Learning While Evaluating (\method).} 
        Given a test case $x_t$, the meta-prompt $M_{t-1}$ generates a sample-specific evaluation prompt $P_t$, which the evaluator uses to produce a judgment $y_t$.
        The evaluator then reflects on its decision to produce self-feedback $f_t$, which is incorporated into the meta-prompt to form $M_t$ for subsequent cases. 
        }
        \label{fig:method}
        \vspace{-0.5em}
    \end{figure*}

    \section{Related Work}\label{section:related_work}
    
    \subsection{LLM-as-a-Judge}
    LLM-as-a-judge~\cite{zheng2023judging} has become a standard alternative to human evaluation, supporting improvements in model behavior during training~\cite{glaese2022improving, bai2022training, ouyang2022training, leerlaif, yuan2024self} as well as inference~\cite{madaan2023self, shinn2023reflexion, zhangaflow}.
    Beyond textual settings, VLM-as-a-judge plays a role in the automatic evaluation of multimodal tasks, including image–text alignment~\cite{chen2024mllm}.

    Pitfalls of LLM-as-a-judge have also been reported, including sensitivity to prompt permutations and various forms of bias, most notably position bias~\cite{zheng2023judging, shi2024judging, koo2024benchmarking, park2024offsetbias, zhao2025one, bavaresco2025llms}. To improve evaluation capability, existing approaches often rely on training~\cite{kim2023prometheus, vu2024foundational, yu2025self, whitehouse2025j1, chan2025j1, lee2024prometheus, zang2025internlm, wang2025unified} or inference-time schemes with multiple models~\cite{jung2025trust, li2025leveraging}.
    Related lines that learn meta-reward models or meta-reasoning procedures~\cite{kim2025evaluativethinkingmetapolicy, saha2025learningplanreason} aim to generate case-specific evaluation criteria or reasoning steps, but they also rely on additional training.
    In contrast, we pursue an inference-only approach that avoids any further training and requires no additional models, making it readily deployable in real-world settings.

    \subsection{Prompt Optimization}
    
    Prompt optimization aims to discover task-specific prompts that enhance model performance~\cite{yang2023large, guo2024connecting, khattab2024dspy, fernandopromptbreeder, wang2024promptagent, yuksekgonul2025optimizing, xumetatextgrad}. 
    These methods iteratively refine prompts using validation sets or external feedback, after which a single optimized prompt is fixed and applied uniformly to all test samples.
    This paradigm is effective, yet inherently \emph{static}: it cannot adapt to sample-specific variations during evaluation.
    Moreover, the reliance on pre-constructed validation sets for optimization can be impractical in real-world evaluation scenarios where constructing labeled data is costly or infeasible.
    To address these limitations, our framework performs meta-prompt evolution without any validation data and generates sample-specific prompts based on the evolving meta-prompt.

    \subsection{Test-Time Scaling}
    
    \paragraph{Per-instance test-time scaling.}
    Recent studies report that allocating more computation at inference time can substantially enhance reasoning quality by increasing deliberation depth, exploring multiple sampled traces, or performing structured search~\cite{snell2024scalingllmtesttimecompute, muennighoff2025s1, shen2025thinking, wang2025mctsjudgetesttimescalingllmasajudge}. 
    However, these approaches operate on a \emph{per-instance} basis: once an instance is completed, its intermediate reasoning is discarded, limiting the model's ability to leverage accumulated experience across test cases.
    
    \paragraph{Sequential test-time scaling.}
    Beyond per-instance compute, several works explore how models can \emph{learn across} test cases by accumulating and reusing experience during inference time~\cite{wang2024agent, liu2025contextual, chen2025log, suzgun2025dynamic, huang2025r2d2}.
    These methods show that models can refine and reuse task-specific heuristics to improve future predictions.
    We instantiate this idea for LLM-as-a-judge through an evolving meta-prompt that stores insights for tailoring sample-specific evaluation instructions and refines itself via self-feedback.
    To further improve sequential methods, our approach allocates inference-time computation \emph{selectively}, guided by a label-free signal (order-swap inconsistency), achieving stronger cost-effectiveness without sacrificing accuracy.

    \begin{algorithm}[t]
        \caption{Learning While Evaluating (\method)}
        \label{alg:lwe}
        \textbf{Input:} A base LLM \texttt{LLM}, a test set $D_{\text{test}} = \{x_t\}_{t=1}^{T}$, an initial meta-prompt $M_0$ and batch size $b$  \\
        \textbf{Output:} A set of evaluated results $S$ and a final meta-prompt $M$
        \begin{algorithmic}[1]
            \STATE $M \leftarrow M_0$ \hfill $\triangleright$ Initialize a meta-prompt.
            \STATE $S \leftarrow \emptyset$ \hfill $\triangleright$ Initialize a set of evaluated results.
            \STATE $F \leftarrow \emptyset$ \hfill $\triangleright$ Buffer for feedback within a batch.
            \FOR{$t = 1$ to $T$}
                \STATE $x_t \leftarrow D_{\text{test}}[t]$
                \STATE $P_t \leftarrow \texttt{BuildEvalPrompt}_{\texttt{LLM}}(M, x_t)$
                \STATE $y_t \leftarrow \texttt{Judge}_{\texttt{LLM}}(P_t, x_t)$
                \STATE $f_t \leftarrow \texttt{Feedback}_{\texttt{LLM}}(M, P_t, x_t, y_t)$
                \STATE $S \leftarrow S \cup \{(x_t, y_t)\}$
                \STATE $F \leftarrow F \cup \{f_t\}$ \hfill 

                \IF{$|F| = b$ \OR $t = T$}
                    \STATE $M \leftarrow \texttt{RefineMetaPrompt}_{\texttt{LLM}}(M, F)$
                    \hfill $\triangleright$ Update the meta-prompt using batch feedback.
                    \STATE $F \leftarrow \emptyset$
                \ENDIF
            \ENDFOR
            \RETURN $S, M$
        \end{algorithmic}
    \end{algorithm}

    \begin{algorithm}[t]
        \caption{\textit{Selective} Learning While Evaluating}
        \label{alg:selective_lwe}
        \textbf{Input:} A base LLM $\texttt{LLM}$, a test set $D_{\text{test}}$, a vanilla prompt $P$, an initial meta-prompt $M_0$ and batch size $b$ \\
        \textbf{Output:} A set of evaluated results $S$ and a final meta-prompt $M$
        \begin{algorithmic}[1]
            \STATE $S \leftarrow \emptyset$ \hfill $\triangleright$ Initialize a set of evaluation results.
            \STATE $I \leftarrow \emptyset$  $\triangleright$ Initialize a set of inconsistent cases.
            \FOR{$x \in D_{\text{test}}$ }
                \STATE $y^{(AB)} \leftarrow \texttt{Judge}_{\texttt{LLM}}(P, x)$ 
                \STATE $x' \leftarrow$ $x$ with the response order swapped
                \STATE $y^{(BA)} \leftarrow \texttt{Judge}_{\texttt{LLM}}(P, x')$
                \IF{$y^{(AB)} = y^{(BA)}$}
                    \STATE $S \leftarrow S \cup \{(x, y^{(AB)})\}$
                     $\triangleright$ Consistent case; skip the update.
                \ELSE
                    \STATE $I \leftarrow I \cup \{x\}$
                     $\triangleright$ Collect inconsistent cases.
                \ENDIF
            \ENDFOR
            \STATE $(S', M) \leftarrow \texttt{LWE}(\texttt{LLM}, I, M_0, b)$
            \STATE $S \leftarrow S \cup S'$
            \RETURN $S, M$
        \end{algorithmic}
    \end{algorithm}   

    \section{Approach}
    We propose that evaluators could benefit from insights accumulated across test cases.  
    In this work, we focus on the pairwise evaluation setup, where the judge compares two candidate responses and identifies the better one.  
    To this end, we introduce Learning While Evaluating (\method), a test-time learning framework that allows evaluators to accumulate and draw on experience from earlier cases during evaluation, ultimately improving their decision-making.

    \subsection{Learning While Evaluating}
    \label{sec:lwe}
    Figure~\ref{fig:method} and Algorithm~\ref{alg:lwe} outline the overall procedure of \method.
    Unlike traditional approaches that evaluate each sample independently, \method processes samples \emph{sequentially}, accumulating insights from previous judgments. 
    It maintains an evolving meta-prompt that (i) produces sample-specific evaluation prompts and (ii) integrates self-generated feedback to improve subsequent evaluations.

    \vspace{0.5em}
    \noindent \textbf{Sequential meta-prompt evolution.}
    Let $\mathcal{D} = \{x_1, x_2, \dots, x_T\}$ denote a set of pairwise comparison cases. 
    Starting from an initial meta-prompt $M_0$, the evaluator sequentially updates its meta-prompt based on observed test cases.
    For each test sample $x_t$, the current meta-prompt $M$ generates a sample-specific prompt $P_t$ (\texttt{BuildEvalPrompt}) to produce a judgment $y_t$ (selecting the better response) (\texttt{Judge}). 
    The evaluator then reflects on this decision to generate feedback $f_t$ (\texttt{Feedback}) for refining $M$ (\texttt{RefineMetaPrompt}). These steps are all realized via LLM prompting.

    To prevent overfitting to individual samples, we update the meta-prompt once per batch of $b$ samples (with $b=4$ in our experiments).
    We analyze the effect of different batch sizes in Figure~\ref{fig:bsz_ablation}.
    Through this sequential process, the meta-prompt captures transferable evaluation heuristics and progressively refines its evaluation strategy across the test set.

    \subsection{\emph{Selective} Learning While Evaluating}
    \label{sec:select_lwe}
    While sequential updates allow leveraging experience across samples, updating the meta-prompt on \textit{every} sample incurs non-trivial computational overhead.
    Moreover, not all samples require the same level of deliberation—straightforward comparisons are already judged accurately by the vanilla evaluator and thus gain little from additional information, whereas challenging cases demand extra guidance for reliable evaluation.
        
    Motivated by prior observations that LLM judges are sensitive to position bias~\cite{zheng2023judging, shi2024judging, koo2024benchmarking}, we turn this vulnerability into a \emph{test-time signal}: we treat order-swap \emph{inconsistency} (\ie{}, disagreement between A\,$\vs$\,B and B\,$\vs$\,A judgments) as a label-free proxy for evaluator uncertainty.

    This choice is attractive for three reasons: (i) it is \emph{available at test time} without ground-truth labels; (ii) it is \emph{instance-specific}, flagging precisely those cases where the current evaluation policy is confused; and (iii) it \emph{targets compute} to the examples that most need additional guidance, avoiding wasted updates on straightforward cases that already yield consistent decisions.
    
    Concretely, as formalized in Algorithm~\ref{alg:selective_lwe}, for each sample $x$, the evaluator makes two vanilla judgments with the response order swapped, and an update is triggered only if the two judgments disagree.
    This selective mechanism bypasses samples with consistent judgments and focuses computational resources on ambiguous cases, improving efficiency without sacrificing accuracy, as we show empirically in Sec.~\ref{sec:results}.

    \section{Experimental Setup}
    \label{sec:exp}

    \subsection{Baselines}
    We compare our proposed methods \method and \methodselective against representative inference approaches.

    \noindent\textbf{Fixed prompting.}
    \texttt{Vanilla} applies a fixed prompt for all test cases.
    \texttt{Chain-of-Thought (CoT)}~\cite{wei2022chain} augments the vanilla prompt with an explicit “step-by-step reasoning” instruction.
    \texttt{Majority Voting} takes the majority label from five independent stochastic judgments of the vanilla prompt (temperature~0.7).
    \texttt{TextGrad}~\citep{yuksekgonul2025optimizing} iteratively modifies its prompt based on validation sets.
    The optimized prompt is then fixed and applied uniformly to all test cases.
    Among all baselines, it is the only method that requires additional labeled data prior to test-time evaluation.\\
    
    \vspace{-1em}
    \noindent\textbf{Adaptive prompting.}
    \texttt{Dynamic Cheatsheet}~\citep{suzgun2025dynamic} (DC) maintains a memory prompt that is sequentially updated during test-time. We use the strongest variant, \texttt{DC-RS}, which retrieves similar examples and curates a memory prompt that conditions the model’s response generation. 
    It is a strong baseline but does not explicitly generate sample-specific evaluation prompts, and it performs updates on every sample unconditionally.
    \texttt{Sample-Specific Prompt} generates a tailored evaluation prompt for each test case from a fixed meta-prompt. 
    Although it applies different prompts per sample, its fixed meta-prompt configuration provides a baseline that isolates the effect of sequence-level meta-prompt updates in our method.

    We use \texttt{gpt-4.1-2025-04-14}~\cite{openai_gpt-4.1-2025-04-14} (gpt-4.1) as a base evaluator for experiments.

    \vspace{0.5em}
    \subsection{Benchmarks}
    We evaluate our method on two multimodal pairwise comparison benchmarks, where the evaluator selects the better response between two candidates.

    \textbf{VL-RewardBench} (VLRewardBench, VL)~\cite{li2025vlrewardbench} and \textbf{Multimodal RewardBench} (MMRewardBench, MM)~\cite{yasunaga2025multimodalrewardbenchholisticevaluation} each provide an image, a question, and two textual responses, and the evaluator must determine which response is better.
    Following prior work~\cite{suzgun2025dynamic}, we use representative subsets where necessary due to API budget constraints: we evaluate on the full 1,247 examples of VLRewardBench, and 1,000 out of 4,711 examples from MMRewardBench.

    \vspace{0.5em}
    \subsection{Evaluation Metrics}
    We consider four metrics. 
    \texttt{Accuracy} is measured on a single random ordering of the two responses. It reflects single-pass performance without accounting for position-bias effects.
    \texttt{Consistency} measures stability under response-order swapping.
    \texttt{Pair Accuracy}, a stricter metric than Accuracy, requires both the original and swapped predictions to be correct and therefore reflects robustness to position bias. 
    This metric is particularly important for assessing the reliability of judge models, as it captures whether their decisions remain stable under input permutations.
    \texttt{Relative Inference Cost} computes the total character length of input and output normalized to the vanilla baseline, providing a quality–cost trade-off measure.
    All metrics are macro-averaged across benchmarks.

    \begin{table*}[t]
    \centering
    \resizebox{\textwidth}{!}{%
    \begin{tabular}{lccccccc}
    \hline
    \multirow{2}{*}{\textbf{Method}} &
    \multicolumn{3}{c}{\textbf{VLRewardBench}} &
    \multicolumn{3}{c}{\textbf{MMRewardBench}} &
    \textbf{Relative Inference Cost}  \\
    \cline{2-7}
     & Acc. ($\uparrow$) & Cons. ($\uparrow$) & PairAcc. ($\uparrow$) 
     & Acc. ($\uparrow$)  & Cons. ($\uparrow$) & PairAcc. ($\uparrow$) 
     & Input \& Output Text ($\downarrow$) \\
    \hline
    Vanilla 
        & 0.629 & 0.801 & 0.529 
        & 0.808 & 0.863 & 0.747 
        & $1.0\times$ \\
    CoT 
        & 0.651 & 0.808 & 0.553
        & 0.808 & 0.874 & 0.749
        & $1.2\times$\\
    Majority Voting 
        & 0.627 & 0.810 & 0.537 
        & 0.828 & 0.891 & 0.769 
        & $5.0\times$ \\
    \hline
    TextGrad\textsuperscript{*}
        & 0.730 & 0.749 & 0.615
        & 0.821 & 0.836 & 0.741
        & $4.4\times$ \\
    \hline
    Dynamic Cheatsheet 
        & 0.698 & 0.868 & 0.629 
        & 0.811 & 0.901 & 0.764 
        & $12.9\times$ \\
    Sample-Specific Prompt
        & 0.661 & 0.727 & 0.529 
        & 0.815 & 0.865 & 0.742 
        & $2.5\times$ \\
    \hline
    \rowcolor{blue!5} \method (\textbf{Ours}) 
        & \textbf{0.745} & 0.805 & 0.646
        & 0.799 & 0.846 & 0.727 
        & $10.9\times$ \\
    \rowcolor{blue!10} \textit{Selective} \method (\textbf{Ours}) 
        & 0.676 & \textbf{0.940} & \textbf{0.648} 
        & \textbf{0.836} & \textbf{0.947} & \textbf{0.808}
        & $3.9\times$ \\
    \hline
    \end{tabular}}
        \caption{
            \textbf{Performance of various inference strategies across benchmarks.}
            We evaluate two vision–language benchmarks (VLRewardBench and MMRewardBench), reporting accuracy (Acc.), consistency (Cons.), and pair accuracy (PairAcc.).
            The rightmost column shows the relative inference cost, measured by total input and output character length normalized to the vanilla baseline.
            Since \method and \methodselective are inherently order-sensitive, 
            we report averaged results over three random-order runs (see Appendix~\ref{appendix:ordering_results_all} for full results).
            Bold indicates the best result among methods that operate without external supervision.
            TextGrad\textsuperscript{*} is included for reference; it relies on gold-labeled validation data and thus operates under a more favorable setting than the other methods (see Appendix~\ref{appendix:textgrad}).
        }

    \label{tab:table1}
    \end{table*}

    \begin{figure*}[t]
        \centering
            \includegraphics[width=1\textwidth]{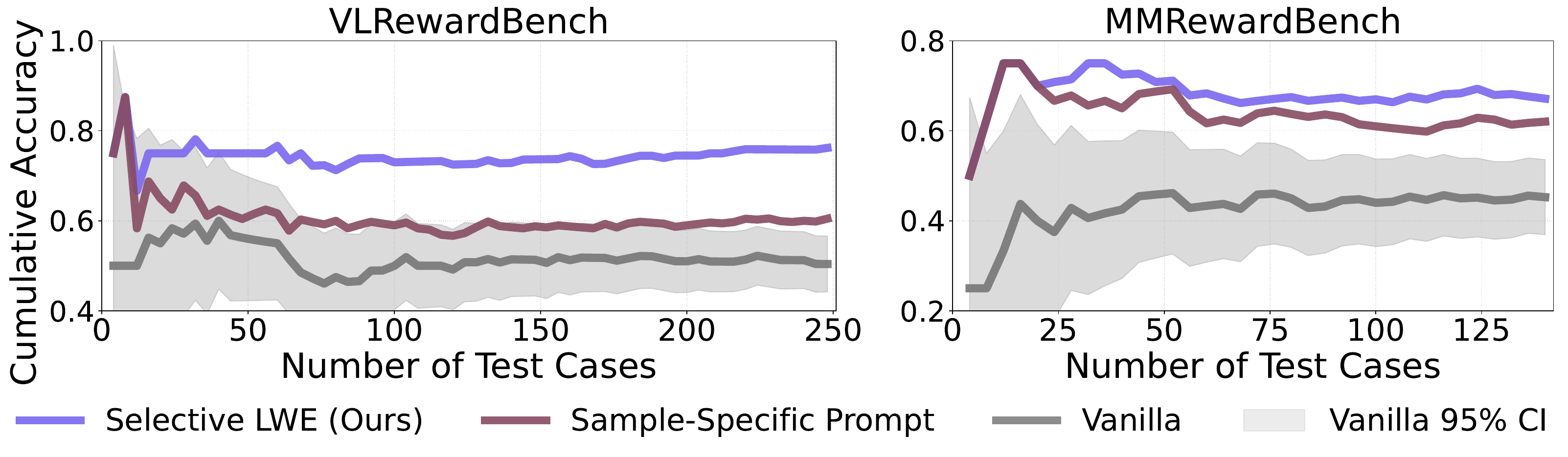}
        \caption{
            \textbf{Cumulative accuracy over test cases.}
            Curves are computed on the vanilla-inconsistent subsets of each benchmark, where \methodselective performs updates.
            \methodselective maintains consistently higher accuracy than both the vanilla baseline and the Sample-Specific Prompt baseline as evaluation progresses, illustrating the benefits of sequential learning and the model’s capacity to integrate insights acquired during evaluation.
            Gray-shaded areas indicate confidence intervals for the vanilla baseline, computed at each point using the binomial proportion method with significance level $\alpha = 0.05$.
        }
        \label{fig:lwe_progression}
        \vspace{-3mm}
    \end{figure*}

    \section{Results}~\label{sec:results}
    Across our experiments, \methodselective consistently improves evaluation quality while operating at substantially lower inference cost than existing adaptive methods.

    \begin{figure}[t!]
        \centering        
        \includegraphics[width=\columnwidth]{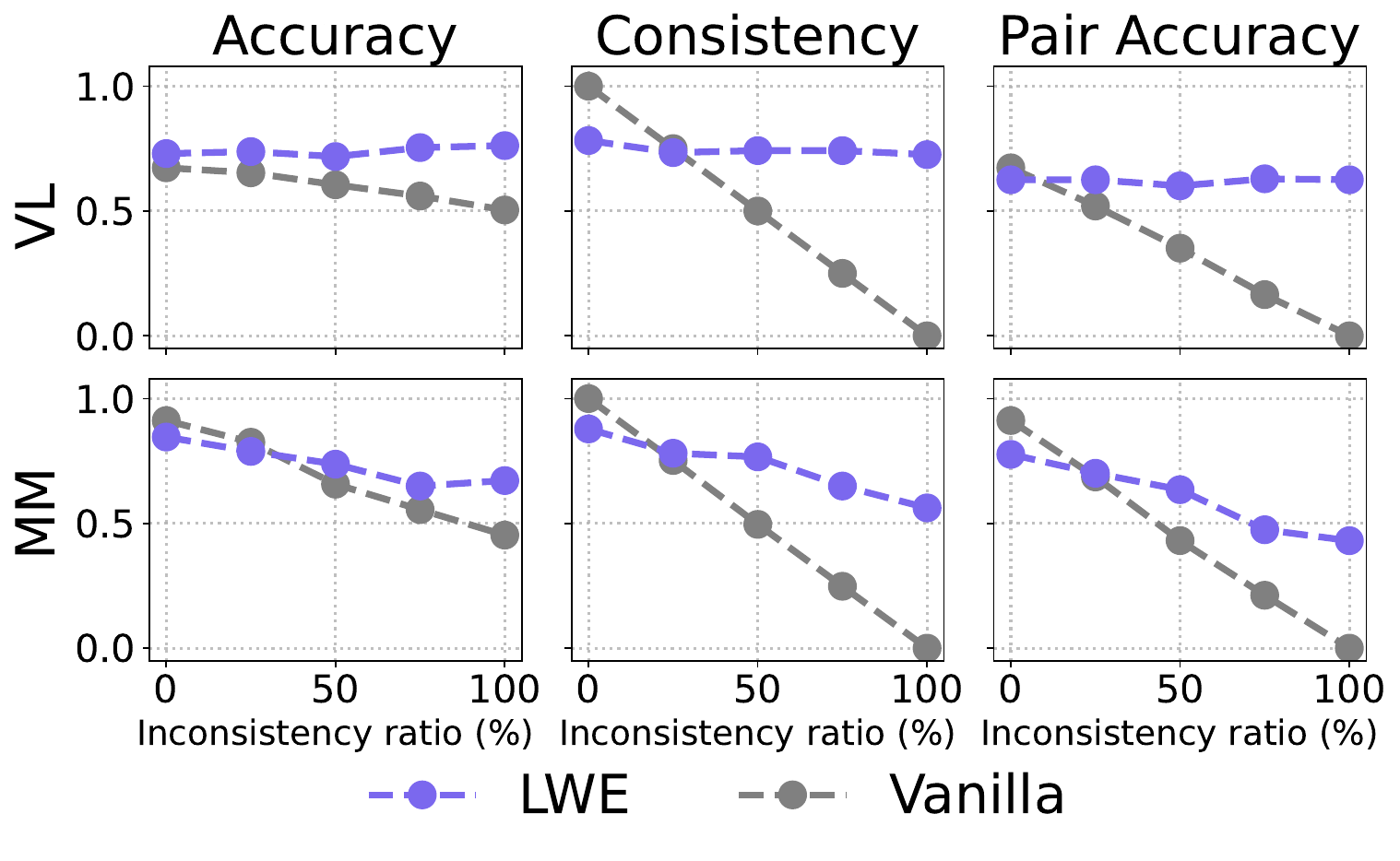}
        \caption{
            \textbf{Effect of inconsistency ratio on meta-prompt updates.}
            We evaluate \method on subsets containing different proportions of inconsistent samples, where inconsistency is computed based on vanilla predictions. 
            As the inconsistency ratio increases, \method shows larger performance gains over the vanilla baseline (purple \vs gray), highlighting its effectiveness in handling inconsistent cases. 
            For each benchmark, the total number of samples is fixed: 248 for VLRewardBench and 137 for MMRewardBench.
        }
        \label{fig:consistency_ratio}
    \end{figure}

    \begin{figure}[t]
            \centering
            \begin{tcolorbox}[colback=blue!5, colframe=blue!50, 
                              boxrule=0.5pt, arc=2mm, left=4pt, right=4pt, top=3pt, bottom=3pt, fontupper=\small]
    
                \noindent
                \dots **Additional Tips:**\\
                - Instruct evaluators to \textcolor{red}{systematically check for calculation or logical errors in each answer, not just the reasoning structure} \textcolor{blue}{penalize both overstatements (e.g., exaggerating color, quantity, or significance) and omissions (e.g., failing to mention a visible but relevant feature)}, and to \textcolor{red}{penalize answers with such errors even if the explanation appears logical.}
                \textcolor{blue}{weigh the impact of these errors on the overall evaluation}.
                \dots \\
                - Require evaluators to ensure that their final justification is \textcolor{red}{comprehensive,} \textcolor{blue}{not only comprehensive but also balanced,} addressing \textcolor{blue}{both the strengths and weaknesses of each answer in relation to} every evaluation criterion \textcolor{red}{ and step, and not omitting any relevant aspect of the comparison}.
            \end{tcolorbox}
            \caption{
                \textbf{Illustration of how the meta-prompt evolves after a single refinement step.}
                Red and blue text denotes instructions \textcolor{red}{removed} and \textcolor{blue}{added} during the update, reflecting the shift from loosely specified checks to clearer, more structured heuristics under \method.
                The full meta-prompts are provided in Appendix~\ref{appendix:fig7_full_prompts}.
            }
            \label{fig:meta_update_example}
        \end{figure}

    \paragraph{Main results.}
    Table~\ref{tab:table1} presents the overall results on the two benchmarks using gpt-4.1.
    \methodselective not only matches or exceeds most baselines in accuracy but also achieves the highest consistency and pair accuracy across all benchmarks, operating at a substantially lower inference cost.
    Notably, \methodselective delivers these gains with only $3.9\times$ relative inference cost, compared to higher costs for other adaptive methods, \ie, $4.4\times$ for TextGrad and $12.9\times$ for DC.

    \paragraph{Learning during evaluation.}
    Figure~\ref{fig:lwe_progression} visualizes the cumulative accuracy on the vanilla-inconsistent subsets of the two benchmarks.
    First, the Sample-Specific Prompt already outperforms the vanilla baseline, demonstrating the benefit of tailoring the evaluation criteria to each case. 
    Building on this, \methodselective achieves even larger gains.
    These results suggest that, as \methodselective progressively refines its meta-prompt through targeted updates, it accumulates evaluation heuristics that extend beyond individual instances, yielding overall improvements in accuracy throughout the evaluation trajectory.

    \paragraph{Full updates vs. \textit{Selective} updates.}
    As shown in Table~\ref{tab:table1}, the inference strategies with updating for all test cases, \ie{}, DC and \method, generally improve all three metrics but require substantially higher inference costs, $12.9\times$ and $10.9\times$ respectively, compared to the vanilla method.
    In contrast, \methodselective preserves most of these gains, often even surpassing the full sequential variant, while operating at only $3.9\times$ the inference cost (approximately 36\% of \method).
    This $3.9\times$ budget reflects two vanilla passes for consistency checking and an additional $1.9\times$ for the LWE process.
    This demonstrates that not every test sample requires equal inference effort: focusing updates on confusing cases provides a more cost-effective path to improved evaluation quality, achieving superior performance and efficiency compared to uniform test-time scaling that updates every sample indiscriminately.
    
    \paragraph{Reliability of judgments.}
    By focusing updates on inconsistent cases, \methodselective gains a substantial increase in consistency (up to 0.947), while also yielding a large margin of pair accuracy.
    This improvement in pair accuracy is particularly meaningful, as it reflects position-invariant evaluation behavior. 
    In real-world evaluation, judges typically assess responses in a random order, so higher pair accuracy implies that such one-shot judgments are more \textit{reliable}.

    \section{Analysis}
    We further analyze the mechanisms underlying the effectiveness of \methodselective, examining whether its selective sequential updates identify informative cases and how robust the resulting behavior is across settings.
    
    \paragraph{Validity of the selection signal.}
    We first examine whether the inconsistency-based signal reliably identifies beneficial samples for meta-prompt updates.
    As shown in Figure~\ref{fig:consistency_ratio}, the performance gains of \method over the vanilla baseline increase monotonically with the proportion of inconsistent samples.
    Notably, \method maintains stable performance even on subsets where the vanilla evaluator’s pair accuracy is zero.
    This reveals an interesting pattern: the evaluator benefits most from the samples that confuse it, making inconsistency an effective guide for updates.
    Consequently, the \textit{Selective} mechanism allocates computation where refinement is most impactful.

    \paragraph{What the meta-prompt learns.}
    To understand how selective updates improve evaluator behavior, we qualitatively examine how the meta-prompt changes during refinement.
    Figure~\ref{fig:meta_update_example} shows that, rather than accumulating ad-hoc advice, updates progressively distill more structured evaluation principles and sharper criteria for distinguishing subtle differences between responses.
    For example, directing the evaluator not only to detect errors but to compare cases where both answers contain issues, assess the impact of overstatements or omissions, and weigh these factors to reach a more balanced and fair judgment.
    This indicates that the meta-prompt is internalizing reliable and portable guidance for future evaluations. See Appendix~\ref{appendix:examples} for evaluation examples.

\begin{figure}[t]
        \centering        
        \includegraphics[width=\columnwidth]{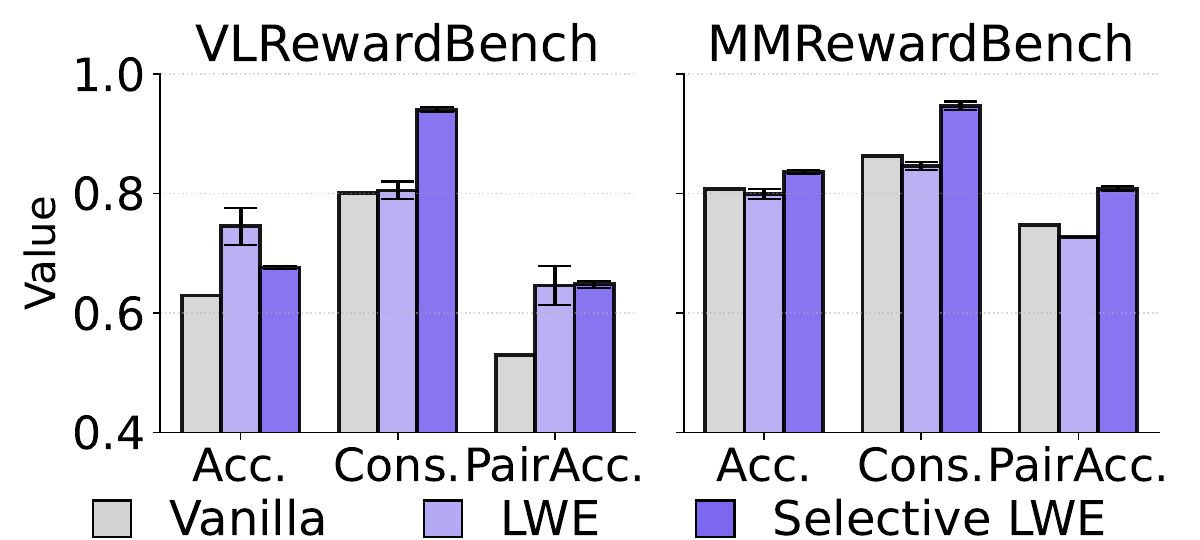}
        \caption{
            \textbf{Ablation on test case ordering.} 
            We evaluate the sensitivity of \method to input ordering using three random permutations of each test set.
            Bars report mean ± standard deviation across permutations.
            \methodselective exhibits stable performance across different orderings.
        }
    
        \label{fig:random_orders}
    \end{figure}

    \paragraph{Robustness to sample ordering.}\label{sec:ordering}
    Since LWE performs sequential updates, we test sensitivity to sample ordering by evaluating each method on three random permutations of each test set.
    As shown in Figure~\ref{fig:random_orders}, results remain consistent across three runs with small standard deviations; on MMRewardBench, the variance of pair accuracy for \method is effectively zero and nearly zero for \methodselective.
    This result suggests that learning effects arise from accumulated experience, rather than incidental ordering artifacts.
    The complete results are provided in Appendix~\ref{appendix:ordering_results_all}.

    \begin{table*}[t!]
    \centering
    \resizebox{\textwidth}{!}{%
        \begin{tabular}{llccccccc}
        \hline
        \multirow{2}{*}{\textbf{Model}} &
        \multirow{2}{*}{\textbf{Method}} &
        \multicolumn{3}{c}{\textbf{VLRewardBench}} &
        \multicolumn{3}{c}{\textbf{MMRewardBench}} &
        \textbf{Relative Inference Cost} \\
        \cline{3-9}
        & & Acc. ($\uparrow$) & Cons. ($\uparrow$) & PairAcc. ($\uparrow$) 
        & Acc. ($\uparrow$)  & Cons. ($\uparrow$) & PairAcc. ($\uparrow$) 
        & Input \& Output Text ($\downarrow$) \\
        \hline
        \multirow{2}{*}{\textbf{gemini-2.5-pro}} 
            & Vanilla
                & 0.754 & 0.888 & 0.696 
                & 0.858 & 0.925 & 0.820 
                & $1.0\times$ \\
            & \cellcolor{blue!10} \textit{Selective} \method 
                & \cellcolor{blue!10} \textbf{0.768} & \cellcolor{blue!10} \textbf{0.955} & \cellcolor{blue!10} \textbf{ 0.744} 
                & \cellcolor{blue!10} \textbf{0.865} & \cellcolor{blue!10} \textbf{0.969} & \cellcolor{blue!10} \textbf{0.852 }
                & \cellcolor{blue!10} $3.2\times$ \\
        \hline
        \multirow{2}{*}{\textbf{claude-sonnet-4.5}} 
            & Vanilla 
                & 0.473 & 0.490 & 0.327 
                & 0.540 & 0.482 & 0.419 
                & $1.0\times$ \\
            & \cellcolor{blue!10}\textit{Selective} \method 
                & \cellcolor{blue!10}\textbf{0.703} & \cellcolor{blue!10}\textbf{0.881} & \cellcolor{blue!10}\textbf{0.648}
                & \cellcolor{blue!10}\textbf{0.823} & \cellcolor{blue!10}\textbf{0.879} & \cellcolor{blue!10}\textbf{0.771}
                & \cellcolor{blue!10}$15.2\times$\\
        \hline
        \end{tabular}
    }
    \caption{
        \textbf{Generalization across evaluators.} 
        Results on gemini-2.5-pro and claude-sonnet-4.5 show that \methodselective consistently improves evaluation performance across evaluators.
        The results of \methodselective are averaged over three random-order runs (see Appendix~\ref{appendix:ordering_results_all} for full results).
    }
    \label{tab:gemini_claude}
    \end{table*}

\begin{figure}[t]
        \centering        
        \includegraphics[width=\columnwidth]{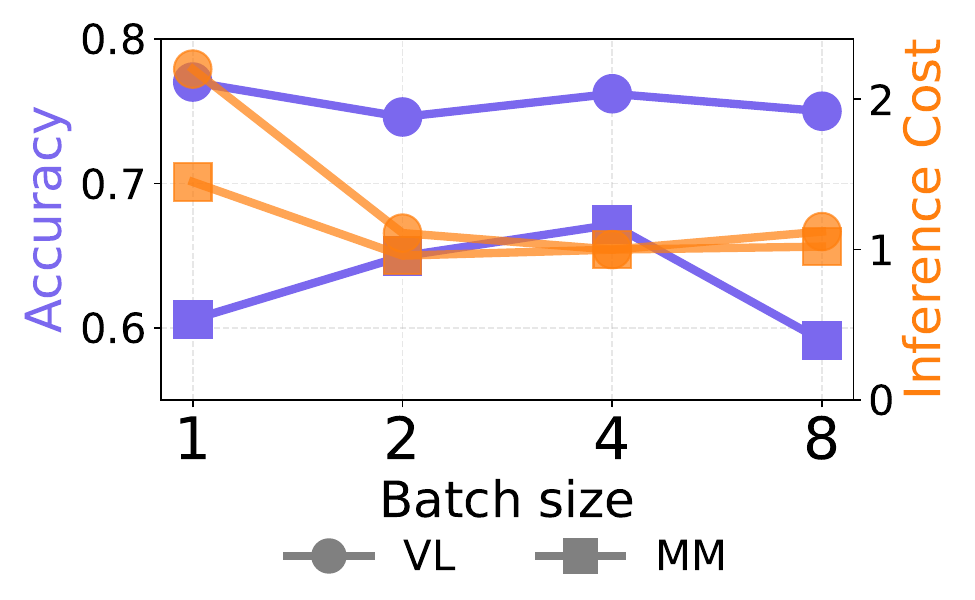}
        \caption{
            \textbf{Effect of batch size ($b$) on accuracy and inference cost.}
            We observe that a moderate batch size ($b{=}4$) yields the best balance between accuracy and inference cost.
            Experiments are conducted on vanilla-inconsistent subsets of each benchmark.
            Inference cost is measured as the ratio of total input and output character counts, normalized to the cost of batch size 4.          
        }
        \label{fig:bsz_ablation}
    \end{figure}

    \paragraph{Effect of batching.}
    We analyze the number of samples processed per update ($b$), as this hyperparameter directly influences two key factors of \method: evaluation performance and inference cost.
    Figure~\ref{fig:bsz_ablation} shows that updating after every sample ($b{=}1$) incurs the highest computational cost, mirroring the overhead of DC, and does not necessarily yield the best performance.
    Conversely, overly large batches ($b{=}8$) degrade performance because the context used for meta-prompt updates becomes excessively long, making the refinement unstable.
    A moderate batch size ($b{=}4$) provides the best balance, sustaining strong accuracy while keeping inference cost relatively low, and we therefore adopt $b{=}4$ as the default configuration of \method.

    \paragraph{Generalization across evaluators.}
    Lastly, we assess whether the benefits of \methodselective extend across different backbone models. 
    As shown in Table~\ref{tab:gemini_claude}, with \texttt{gemini-2.5-pro}~\cite{comanici2025gemini}, which already achieves strong vanilla performance, \methodselective still yields measurable gains, indicating that the selective updates extract additional signal even when the base evaluator is already competent.
    
    On the other hand, \texttt{claude-sonnet-4.5}~\cite{anthropic2025claude45} shows substantially larger improvements.
    This is explained by \methodselective’s strength on vanilla-inconsistent cases: claude-sonnet-4.5 produces a large number of such confusing instances (over 50\% of the test sets), allowing the method to intervene more often and correct failures where the vanilla evaluator struggles.
    Its elevated inference cost ($15.2\times$) naturally follows from this higher frequency of selective updates. 
    
    Despite these model-specific differences, the results reveal a consistent pattern of improvement, positioning our method as a readily deployable inference strategy across diverse generative evaluators.

    \section{Conclusion}
    In this work, we study sequential test-time learning for evaluators and ask whether they can \emph{learn while testing}.
    We introduce \textbf{Learning While Evaluating} (\method), a framework that maintains an evolving meta-prompt to generate sample-specific evaluation instructions and refine itself via self-generated feedback.
    We further propose \methodselective, which updates only on samples exhibiting self-inconsistency.
    Across two pairwise comparison benchmarks, we show that \methodselective achieves strong evaluation performance at only a fraction of the token cost required by full sequential updates.
    These results demonstrate that focusing compute on confusing cases yields reliable and cost-efficient evaluation, without training or external supervision.
    We hope our work encourages a shift from static judges that rely on a single fixed prompt for all samples toward \emph{learning evaluators} that improve \emph{during} evaluation—accumulating experience as they progress through test cases, tailoring sample-specific criteria, and delivering more reliable judgments within practical inference budgets.

    \section*{Limitations}
    Our method relies on the base capability of the underlying model, making it less effective for relatively weak models.
    When applied to the open model \texttt{Qwen3-VL-235B-A22B-Instruct}~\cite{qwen3vl2025}, which achieves competent vanilla performance, the model often fails to produce valid evaluation prompts under meta-prompt updates.
    
    We perform two rounds of inference over the entire test set and leverage internal inconsistency to identify cases for updates.
    However, some “consistent but wrong” cases remain indistinguishable without access to ground-truth labels, which we leave for future investigation.
    
    Finally, while our study focuses on pairwise comparisons, the proposed \method can extend to direct assessment and other evaluation settings.

    \section*{Acknowledgments}
    We thank the members of SNUMPR, especially Seongwon Cho, San Kim, and Hyeonbeom Choi, for their valuable comments and support.

    \newpage

\bibliography{custom}

\begin{thebibliography}{54}
\providecommand{\natexlab}[1]{#1}

\bibitem[{{Anthropic}(2025)}]{anthropic2025claude45}
{Anthropic}. 2025.
\newblock \href {https://assets.anthropic.com/m/12f214efcc2f457a/original/Claude-Sonnet-4-5-System-Card.pdf} {Claude sonnet 4.5 system card}.
\newblock Technical report, Anthropic PBC.
\newblock Accessed: 2025-11-17.

\bibitem[{Bai et~al.(2022)Bai, Jones, Ndousse, Askell, Chen, DasSarma, Drain, Fort, Ganguli, Henighan et~al.}]{bai2022training}
Yuntao Bai, Andy Jones, Kamal Ndousse, Amanda Askell, Anna Chen, Nova DasSarma, Dawn Drain, Stanislav Fort, Deep Ganguli, Tom Henighan, and 1 others. 2022.
\newblock Training a helpful and harmless assistant with reinforcement learning from human feedback.
\newblock \emph{arXiv preprint arXiv:2204.05862}.

\bibitem[{Bavaresco et~al.(2025)Bavaresco, Bernardi, Bertolazzi, Elliott, Fern{\'a}ndez, Gatt, Ghaleb, Giulianelli, Hanna, Koller et~al.}]{bavaresco2025llms}
Anna Bavaresco, Raffaella Bernardi, Leonardo Bertolazzi, Desmond Elliott, Raquel Fern{\'a}ndez, Albert Gatt, Esam Ghaleb, Mario Giulianelli, Michael Hanna, Alexander Koller, and 1 others. 2025.
\newblock Llms instead of human judges? a large scale empirical study across 20 nlp evaluation tasks.
\newblock In \emph{Proceedings of the 63rd Annual Meeting of the Association for Computational Linguistics (Volume 2: Short Papers)}, pages 238--255.

\bibitem[{Chan et~al.(2025)Chan, Xu, Ji, Ye, Wen, Jiang, Yang, Xue, Han, and Guo}]{chan2025j1}
Chi-Min Chan, Chunpu Xu, Jiaming Ji, Zhen Ye, Pengcheng Wen, Chunyang Jiang, Yaodong Yang, Wei Xue, Sirui Han, and Yike Guo. 2025.
\newblock J1: Exploring simple test-time scaling for llm-as-a-judge.
\newblock \emph{arXiv preprint arXiv:2505.11875}.

\bibitem[{Chen et~al.(2024)Chen, Chen, Zhang, Wang, Liu, Zhou, Zhang, Wan, Zhou, and Sun}]{chen2024mllm}
Dongping Chen, Ruoxi Chen, Shilin Zhang, Yaochen Wang, Yinuo Liu, Huichi Zhou, Qihui Zhang, Yao Wan, Pan Zhou, and Lichao Sun. 2024.
\newblock Mllm-as-a-judge: Assessing multimodal llm-as-a-judge with vision-language benchmark.
\newblock In \emph{Forty-first International Conference on Machine Learning}.

\bibitem[{Chen et~al.(2025)Chen, Zhang, Roth, Madden, Andreas, and Cafarella}]{chen2025log}
Peter~Baile Chen, Yi~Zhang, Dan Roth, Samuel Madden, Jacob Andreas, and Michael Cafarella. 2025.
\newblock Log-augmented generation: Scaling test-time reasoning with reusable computation.
\newblock \emph{arXiv preprint arXiv:2505.14398}.

\bibitem[{Comanici et~al.(2025)Comanici, Bieber, Schaekermann, Pasupat, Sachdeva, Dhillon, Blistein, Ram, Zhang, Rosen et~al.}]{comanici2025gemini}
Gheorghe Comanici, Eric Bieber, Mike Schaekermann, Ice Pasupat, Noveen Sachdeva, Inderjit Dhillon, Marcel Blistein, Ori Ram, Dan Zhang, Evan Rosen, and 1 others. 2025.
\newblock Gemini 2.5: Pushing the frontier with advanced reasoning, multimodality, long context, and next generation agentic capabilities.
\newblock \emph{arXiv preprint arXiv:2507.06261}.

\bibitem[{Fernando et~al.(2023)Fernando, Banarse, Michalewski, Osindero, and Rockt{\"a}schel}]{fernandopromptbreeder}
Chrisantha Fernando, Dylan~Sunil Banarse, Henryk Michalewski, Simon Osindero, and Tim Rockt{\"a}schel. 2023.
\newblock Promptbreeder: Self-referential self-improvement via prompt evolution.
\newblock In \emph{Forty-first International Conference on Machine Learning}.

\bibitem[{Flavell(1979)}]{flavell1979metacognition}
John~H. Flavell. 1979.
\newblock \href {https://doi.org/10.1037/0003-066X.34.10.906} {Metacognition and cognitive monitoring: A new area of cognitive-developmental inquiry}.
\newblock \emph{American Psychologist}, 34(10):906--911.

\bibitem[{Glaese et~al.(2022)Glaese, McAleese, Tr{\k{e}}bacz, Aslanides, Firoiu, Ewalds, Rauh, Weidinger, Chadwick, Thacker et~al.}]{glaese2022improving}
Amelia Glaese, Nat McAleese, Maja Tr{\k{e}}bacz, John Aslanides, Vlad Firoiu, Timo Ewalds, Maribeth Rauh, Laura Weidinger, Martin Chadwick, Phoebe Thacker, and 1 others. 2022.
\newblock Improving alignment of dialogue agents via targeted human judgements.
\newblock \emph{arXiv preprint arXiv:2209.14375}.

\bibitem[{Grattafiori et~al.(2024)Grattafiori, Dubey, Jauhri, Pandey, Kadian, Al-Dahle, Letman, Mathur, Schelten, Vaughan et~al.}]{grattafiori2024llama}
Aaron Grattafiori, Abhimanyu Dubey, Abhinav Jauhri, Abhinav Pandey, Abhishek Kadian, Ahmad Al-Dahle, Aiesha Letman, Akhil Mathur, Alan Schelten, Alex Vaughan, and 1 others. 2024.
\newblock The llama 3 herd of models.
\newblock \emph{arXiv preprint arXiv:2407.21783}.

\bibitem[{Guo et~al.(2024)Guo, Wang, Guo, Li, Song, Tan, Liu, Bian, and Yang}]{guo2024connecting}
Qingyan Guo, Rui Wang, Junliang Guo, Bei Li, Kaitao Song, Xu~Tan, Guoqing Liu, Jiang Bian, and Yujiu Yang. 2024.
\newblock \href {https://openreview.net/forum?id=ZG3RaNIsO8} {Connecting large language models with evolutionary algorithms yields powerful prompt optimizers}.
\newblock In \emph{The Twelfth International Conference on Learning Representations}.

\bibitem[{Huang et~al.(2025)Huang, Basu, Abdelaziz, Kapanipathi, May, and Chen}]{huang2025r2d2}
Tenghao Huang, Kinjal Basu, Ibrahim Abdelaziz, Pavan Kapanipathi, Jonathan May, and Muhao Chen. 2025.
\newblock R2d2: Remembering, replaying and dynamic decision making with a reflective agentic memory.
\newblock In \emph{Proceedings of the 63rd Annual Meeting of the Association for Computational Linguistics (Volume 1: Long Papers)}, pages 30318--30330.

\bibitem[{Jung et~al.(2025)Jung, Brahman, and Choi}]{jung2025trust}
Jaehun Jung, Faeze Brahman, and Yejin Choi. 2025.
\newblock \href {https://openreview.net/forum?id=UHPnqSTBPO} {Trust or escalate: {LLM} judges with provable guarantees for human agreement}.
\newblock In \emph{The Thirteenth International Conference on Learning Representations}.

\bibitem[{Khattab et~al.(2024)Khattab, Singhvi, Maheshwari, Zhang, Santhanam, Haq, Sharma, Joshi, Moazam, Miller et~al.}]{khattab2024dspy}
Omar Khattab, Arnav Singhvi, Paridhi Maheshwari, Zhiyuan Zhang, Keshav Santhanam, Saiful Haq, Ashutosh Sharma, Thomas~T Joshi, Hanna Moazam, Heather Miller, and 1 others. 2024.
\newblock Dspy: Compiling declarative language model calls into state-of-the-art pipelines.
\newblock In \emph{The Twelfth International Conference on Learning Representations}.

\bibitem[{Kim et~al.(2023)Kim, Shin, Cho, Jang, Longpre, Lee, Yun, Shin, Kim, Thorne et~al.}]{kim2023prometheus}
Seungone Kim, Jamin Shin, Yejin Cho, Joel Jang, Shayne Longpre, Hwaran Lee, Sangdoo Yun, Seongjin Shin, Sungdong Kim, James Thorne, and 1 others. 2023.
\newblock Prometheus: Inducing fine-grained evaluation capability in language models.
\newblock In \emph{The Twelfth International Conference on Learning Representations}.

\bibitem[{Kim et~al.(2025)Kim, Park, Raheja, Kim, and Kang}]{kim2025evaluativethinkingmetapolicy}
Zae~Myung Kim, Chanwoo Park, Vipul Raheja, Suin Kim, and Dongyeop Kang. 2025.
\newblock \href {https://arxiv.org/abs/2504.20157} {Toward evaluative thinking: Meta policy optimization with evolving reward models}.
\newblock \emph{Preprint}, arXiv:2504.20157.

\bibitem[{Koo et~al.(2024)Koo, Lee, Raheja, Park, Kim, and Kang}]{koo2024benchmarking}
Ryan Koo, Minhwa Lee, Vipul Raheja, Jong~Inn Park, Zae~Myung Kim, and Dongyeop Kang. 2024.
\newblock Benchmarking cognitive biases in large language models as evaluators.
\newblock In \emph{Findings of the Association for Computational Linguistics ACL 2024}, pages 517--545.

\bibitem[{Lee et~al.(2023)Lee, Phatale, Mansoor, Mesnard, Ferret, Lu, Bishop, Hall, Carbune, Rastogi et~al.}]{leerlaif}
Harrison Lee, Samrat Phatale, Hassan Mansoor, Thomas Mesnard, Johan Ferret, Kellie~Ren Lu, Colton Bishop, Ethan Hall, Victor Carbune, Abhinav Rastogi, and 1 others. 2023.
\newblock Rlaif vs. rlhf: Scaling reinforcement learning from human feedback with ai feedback.
\newblock In \emph{Forty-first International Conference on Machine Learning}.

\bibitem[{Lee et~al.(2024)Lee, Kim, Park, Kim, and Seo}]{lee2024prometheus}
Seongyun Lee, Seungone Kim, Sue Park, Geewook Kim, and Minjoon Seo. 2024.
\newblock Prometheus-vision: Vision-language model as a judge for fine-grained evaluation.
\newblock In \emph{Findings of the association for computational linguistics ACL 2024}, pages 11286--11315.

\bibitem[{Li et~al.(2025{\natexlab{a}})Li, Wei, Xie, Yang, Song, Wang, An, Liu, Li, Lin, Kong, and Liu}]{li2025vlrewardbench}
Lei Li, Yuancheng Wei, Zhihui Xie, Xuqing Yang, Yifan Song, Peiyi Wang, Chenxin An, Tianyu Liu, Sujian Li, Bill~Yuchen Lin, Lingpeng Kong, and Qi~Liu. 2025{\natexlab{a}}.
\newblock Vl-rewardbench: A challenging benchmark for vision-language generative reward models.
\newblock In \emph{{CVPR}}.

\bibitem[{Li et~al.(2025{\natexlab{b}})Li, Mohamud, Sun, Wu, and Boulet}]{li2025leveraging}
Yuran Li, Jama~Hussein Mohamud, Chongren Sun, Di~Wu, and Benoit Boulet. 2025{\natexlab{b}}.
\newblock Leveraging llms as meta-judges: A multi-agent framework for evaluating llm judgments.
\newblock \emph{arXiv preprint arXiv:2504.17087}.

\bibitem[{Liu et~al.(2025)Liu, Si, Narasimhan, and Yao}]{liu2025contextual}
Yitao Liu, Chenglei Si, Karthik Narasimhan, and Shunyu Yao. 2025.
\newblock Contextual experience replay for self-improvement of language agents.
\newblock \emph{arXiv preprint arXiv:2506.06698}.

\bibitem[{Madaan et~al.(2023)Madaan, Tandon, Gupta, Hallinan, Gao, Wiegreffe, Alon, Dziri, Prabhumoye, Yang et~al.}]{madaan2023self}
Aman Madaan, Niket Tandon, Prakhar Gupta, Skyler Hallinan, Luyu Gao, Sarah Wiegreffe, Uri Alon, Nouha Dziri, Shrimai Prabhumoye, Yiming Yang, and 1 others. 2023.
\newblock Self-refine: Iterative refinement with self-feedback.
\newblock \emph{Advances in Neural Information Processing Systems}, 36:46534--46594.

\bibitem[{Muennighoff et~al.(2025)Muennighoff, Yang, Shi, Li, Fei-Fei, Hajishirzi, Zettlemoyer, Liang, Cand{\`e}s, and Hashimoto}]{muennighoff2025s1}
Niklas Muennighoff, Zitong Yang, Weijia Shi, Xiang~Lisa Li, Li~Fei-Fei, Hannaneh Hajishirzi, Luke Zettlemoyer, Percy Liang, Emmanuel Cand{\`e}s, and Tatsunori Hashimoto. 2025.
\newblock s1: Simple test-time scaling.
\newblock \emph{arXiv preprint arXiv:2501.19393}.

\bibitem[{OpenAI(2023)}]{openai-evals}
OpenAI. 2023.
\newblock Openai evals repository.
\newblock \url{https://github.com/openai/evals}.
\newblock Accessed: 2025-10-05.

\bibitem[{{OpenAI}(2025)}]{openai_gpt-4.1-2025-04-14}
{OpenAI}. 2025.
\newblock \href {https://platform.openai.com/docs/models/gpt-4.1} {{GPT-4.1-2025-04-14}}.
\newblock Accessed: 2025-10-05.

\bibitem[{Ouyang et~al.(2022)Ouyang, Wu, Jiang, Almeida, Wainwright, Mishkin, Zhang, Agarwal, Slama, Ray et~al.}]{ouyang2022training}
Long Ouyang, Jeffrey Wu, Xu~Jiang, Diogo Almeida, Carroll Wainwright, Pamela Mishkin, Chong Zhang, Sandhini Agarwal, Katarina Slama, Alex Ray, and 1 others. 2022.
\newblock Trainfing language models to follow instructions with human feedback.
\newblock \emph{Advances in neural information processing systems}, 35:27730--27744.

\bibitem[{Park et~al.(2024)Park, Jwa, Meiying, Kim, and Choi}]{park2024offsetbias}
Junsoo Park, Seungyeon Jwa, Ren Meiying, Daeyoung Kim, and Sanghyuk Choi. 2024.
\newblock Offsetbias: Leveraging debiased data for tuning evaluators.
\newblock In \emph{Findings of the Association for Computational Linguistics: EMNLP 2024}, pages 1043--1067.

\bibitem[{{Qwen team, Alibaba Cloud}(2025)}]{qwen3vl2025}
{Qwen team, Alibaba Cloud}. 2025.
\newblock Qwen3-vl.
\newblock \url{https://github.com/QwenLM/Qwen3-VL}.
\newblock Accessed: 2025-10-05.

\bibitem[{Saha et~al.(2025)Saha, Li, Ghazvininejad, Weston, and Wang}]{saha2025learningplanreason}
Swarnadeep Saha, Xian Li, Marjan Ghazvininejad, Jason Weston, and Tianlu Wang. 2025.
\newblock \href {https://arxiv.org/abs/2501.18099} {Learning to plan \& reason for evaluation with thinking-llm-as-a-judge}.
\newblock \emph{Preprint}, arXiv:2501.18099.

\bibitem[{Shen et~al.(2025)Shen, Bai, Zhang, Zhou, Setlur, Tong, Caples, Jiang, Zhang, Talwalkar et~al.}]{shen2025thinking}
Junhong Shen, Hao Bai, Lunjun Zhang, Yifei Zhou, Amrith Setlur, Shengbang Tong, Diego Caples, Nan Jiang, Tong Zhang, Ameet Talwalkar, and 1 others. 2025.
\newblock Thinking vs. doing: Agents that reason by scaling test-time interaction.
\newblock \emph{arXiv preprint arXiv:2506.07976}.

\bibitem[{Shi et~al.(2024)Shi, Ma, Liang, Diao, Ma, and Vosoughi}]{shi2024judging}
Lin Shi, Chiyu Ma, Wenhua Liang, Xingjian Diao, Weicheng Ma, and Soroush Vosoughi. 2024.
\newblock Judging the judges: A systematic study of position bias in llm-as-a-judge.
\newblock \emph{arXiv preprint arXiv:2406.07791}.

\bibitem[{Shinn et~al.(2023)Shinn, Cassano, Gopinath, Narasimhan, and Yao}]{shinn2023reflexion}
Noah Shinn, Federico Cassano, Ashwin Gopinath, Karthik Narasimhan, and Shunyu Yao. 2023.
\newblock Reflexion: Language agents with verbal reinforcement learning.
\newblock \emph{Advances in Neural Information Processing Systems}, 36:8634--8652.

\bibitem[{Snell et~al.(2024)Snell, Lee, Xu, and Kumar}]{snell2024scalingllmtesttimecompute}
Charlie Snell, Jaehoon Lee, Kelvin Xu, and Aviral Kumar. 2024.
\newblock \href {https://arxiv.org/abs/2408.03314} {Scaling llm test-time compute optimally can be more effective than scaling model parameters}.
\newblock \emph{Preprint}, arXiv:2408.03314.

\bibitem[{Sternberg(1985)}]{sternberg1985beyond}
Robert~J. Sternberg. 1985.
\newblock \emph{Beyond IQ: A triarchic theory of human intelligence}.
\newblock Cambridge University Press, New York.

\bibitem[{Suzgun et~al.(2025)Suzgun, Yuksekgonul, Bianchi, Jurafsky, and Zou}]{suzgun2025dynamic}
Mirac Suzgun, Mert Yuksekgonul, Federico Bianchi, Dan Jurafsky, and James Zou. 2025.
\newblock Dynamic cheatsheet: Test-time learning with adaptive memory.
\newblock \emph{arXiv preprint arXiv:2504.07952}.

\bibitem[{Vu et~al.(2024)Vu, Krishna, Alzubi, Tar, Faruqui, and Sung}]{vu2024foundational}
Tu~Vu, Kalpesh Krishna, Salaheddin Alzubi, Chris Tar, Manaal Faruqui, and Yun-Hsuan Sung. 2024.
\newblock Foundational autoraters: Taming large language models for better automatic evaluation.
\newblock In \emph{Proceedings of the 2024 Conference on Empirical Methods in Natural Language Processing}, pages 17086--17105.

\bibitem[{Wang et~al.(2024{\natexlab{a}})Wang, Li, Wang, Bai, Luo, Zhang, Jojic, Xing, and Hu}]{wang2024promptagent}
Xinyuan Wang, Chenxi Li, Zhen Wang, Fan Bai, Haotian Luo, Jiayou Zhang, Nebojsa Jojic, Eric Xing, and Zhiting Hu. 2024{\natexlab{a}}.
\newblock \href {https://openreview.net/forum?id=22pyNMuIoa} {Promptagent: Strategic planning with language models enables expert-level prompt optimization}.
\newblock In \emph{The Twelfth International Conference on Learning Representations}.

\bibitem[{Wang et~al.(2025{\natexlab{a}})Wang, Zang, Li, Jin, and Wang}]{wang2025unified}
Yibin Wang, Yuhang Zang, Hao Li, Cheng Jin, and Jiaqi Wang. 2025{\natexlab{a}}.
\newblock Unified reward model for multimodal understanding and generation.
\newblock \emph{arXiv preprint arXiv:2503.05236}.

\bibitem[{Wang et~al.(2025{\natexlab{b}})Wang, Ji, Yang, Li, Hu, Li, and Sartoretti}]{wang2025mctsjudgetesttimescalingllmasajudge}
Yutong Wang, Pengliang Ji, Chaoqun Yang, Kaixin Li, Ming Hu, Jiaoyang Li, and Guillaume Sartoretti. 2025{\natexlab{b}}.
\newblock \href {https://arxiv.org/abs/2502.12468} {Mcts-judge: Test-time scaling in llm-as-a-judge for code correctness evaluation}.
\newblock \emph{Preprint}, arXiv:2502.12468.

\bibitem[{Wang et~al.(2024{\natexlab{b}})Wang, Mao, Fried, and Neubig}]{wang2024agent}
Zora~Zhiruo Wang, Jiayuan Mao, Daniel Fried, and Graham Neubig. 2024{\natexlab{b}}.
\newblock Agent workflow memory.
\newblock \emph{arXiv preprint arXiv:2409.07429}.

\bibitem[{Wei et~al.(2022)Wei, Wang, Schuurmans, Bosma, brian ichter, Xia, Chi, Le, and Zhou}]{wei2022chain}
Jason Wei, Xuezhi Wang, Dale Schuurmans, Maarten Bosma, brian ichter, Fei Xia, Ed~H. Chi, Quoc~V Le, and Denny Zhou. 2022.
\newblock \href {https://openreview.net/forum?id=_VjQlMeSB_J} {Chain of thought prompting elicits reasoning in large language models}.
\newblock In \emph{Advances in Neural Information Processing Systems}.

\bibitem[{Whitehouse et~al.(2025)Whitehouse, Wang, Yu, Li, Weston, Kulikov, and Saha}]{whitehouse2025j1}
Chenxi Whitehouse, Tianlu Wang, Ping Yu, Xian Li, Jason Weston, Ilia Kulikov, and Swarnadeep Saha. 2025.
\newblock J1: Incentivizing thinking in llm-as-a-judge via reinforcement learning.
\newblock \emph{arXiv preprint arXiv:2505.10320}.

\bibitem[{Xu et~al.(2025)Xu, Yuksekgonul, Guestrin, and Zou}]{xumetatextgrad}
Guowei Xu, Mert Yuksekgonul, Carlos Guestrin, and James Zou. 2025.
\newblock metatextgrad: Learning to learn with language models as optimizers.
\newblock In \emph{Adaptive Foundation Models: Evolving AI for Personalized and Efficient Learning}.

\bibitem[{Yang et~al.(2023)Yang, Wang, Lu, Liu, Le, Zhou, and Chen}]{yang2023large}
Chengrun Yang, Xuezhi Wang, Yifeng Lu, Hanxiao Liu, Quoc~V Le, Denny Zhou, and Xinyun Chen. 2023.
\newblock Large language models as optimizers.
\newblock In \emph{The Twelfth International Conference on Learning Representations}.

\bibitem[{Yasunaga et~al.(2025)Yasunaga, Zettlemoyer, and Ghazvininejad}]{yasunaga2025multimodalrewardbenchholisticevaluation}
Michihiro Yasunaga, Luke Zettlemoyer, and Marjan Ghazvininejad. 2025.
\newblock \href {https://arxiv.org/abs/2502.14191} {Multimodal rewardbench: Holistic evaluation of reward models for vision language models}.
\newblock \emph{Preprint}, arXiv:2502.14191.

\bibitem[{Yu et~al.(2025)Yu, Chen, Zhang, Tan, Zhu, Pang, Qian, Wang, Gururangan, Zhang et~al.}]{yu2025self}
Yue Yu, Zhengxing Chen, Aston Zhang, Liang Tan, Chenguang Zhu, Richard~Yuanzhe Pang, Yundi Qian, Xuewei Wang, Suchin Gururangan, Chao Zhang, and 1 others. 2025.
\newblock Self-generated critiques boost reward modeling for language models.
\newblock In \emph{Proceedings of the 2025 Conference of the Nations of the Americas Chapter of the Association for Computational Linguistics: Human Language Technologies (Volume 1: Long Papers)}, pages 11499--11514.

\bibitem[{Yuan et~al.(2024)Yuan, Pang, Cho, Li, Sukhbaatar, Xu, and Weston}]{yuan2024self}
Weizhe Yuan, Richard~Yuanzhe Pang, Kyunghyun Cho, Xian Li, Sainbayar Sukhbaatar, Jing Xu, and Jason~E Weston. 2024.
\newblock Self-rewarding language models.
\newblock In \emph{Forty-first International Conference on Machine Learning}.

\bibitem[{Yuksekgonul et~al.(2025)Yuksekgonul, Bianchi, Boen, Liu, Lu, Huang, Guestrin, and Zou}]{yuksekgonul2025optimizing}
Mert Yuksekgonul, Federico Bianchi, Joseph Boen, Sheng Liu, Pan Lu, Zhi Huang, Carlos Guestrin, and James Zou. 2025.
\newblock Optimizing generative ai by backpropagating language model feedback.
\newblock \emph{Nature}, 639:609--616.

\bibitem[{Zang et~al.(2025)Zang, Dong, Zhang, Cao, Liu, Ding, Wu, Ma, Duan, Zhang et~al.}]{zang2025internlm}
Yuhang Zang, Xiaoyi Dong, Pan Zhang, Yuhang Cao, Ziyu Liu, Shengyuan Ding, Shenxi Wu, Yubo Ma, Haodong Duan, Wenwei Zhang, and 1 others. 2025.
\newblock Internlm-xcomposer2. 5-reward: A simple yet effective multi-modal reward model.
\newblock \emph{arXiv preprint arXiv:2501.12368}.

\bibitem[{Zhang et~al.(2024)Zhang, Xiang, Yu, Teng, Chen, Chen, Zhuge, Cheng, Hong, Wang et~al.}]{zhangaflow}
Jiayi Zhang, Jinyu Xiang, Zhaoyang Yu, Fengwei Teng, Xiong-Hui Chen, Jiaqi Chen, Mingchen Zhuge, Xin Cheng, Sirui Hong, Jinlin Wang, and 1 others. 2024.
\newblock Aflow: Automating agentic workflow generation.
\newblock In \emph{The Thirteenth International Conference on Learning Representations}.

\bibitem[{Zhao et~al.(2025)Zhao, Liu, Yu, Kung, Chen, Mi, and Yu}]{zhao2025one}
Yulai Zhao, Haolin Liu, Dian Yu, Sunyuan Kung, Meijia Chen, Haitao Mi, and Dong Yu. 2025.
\newblock One token to fool llm-as-a-judge.
\newblock \emph{arXiv preprint arXiv:2507.08794}.

\bibitem[{Zheng et~al.(2023)Zheng, Chiang, Sheng, Zhuang, Wu, Zhuang, Lin, Li, Li, Xing et~al.}]{zheng2023judging}
Lianmin Zheng, Wei-Lin Chiang, Ying Sheng, Siyuan Zhuang, Zhanghao Wu, Yonghao Zhuang, Zi~Lin, Zhuohan Li, Dacheng Li, Eric Xing, and 1 others. 2023.
\newblock Judging llm-as-a-judge with mt-bench and chatbot arena.
\newblock \emph{Advances in neural information processing systems}, 36:46595--46623.

\end{thebibliography}

\appendix

\clearpage

\section*{Appendix}

\section{Implementation Details} \label{subsection:details}
For meta-prompt updates, only one-sided judgments are utilized. The swapped counterparts are excluded to avoid nearly doubling the context length and to maintain inference efficiency.

As evaluation progresses, the meta-prompt accumulates insights and can grow excessively long.
Empirically, we observe that beyond a certain length, additional content becomes redundant and yields only marginal performance gains.
To address this, we periodically summarize the meta-prompt once it exceeds a predefined length threshold (10,000 characters in our experiments).

For Multimodal RewardBench, we subsampled 1,000 cases from the 4,711 samples, excluding the 500 samples from the “Hateful Memes” subset. This subset was included in the original paper but not provided in the benchmark’s official implementation, so we used the data available in the official release.

For a fair comparison, Dynamic Cheatsheet, which is fully sequential, was evaluated using the same test-case order as LWE (run0 in Appendix~\ref{appendix:ordering_results_all}).

All experiments with gpt-4.1, gemini-2.5-pro, and claude-sonnet-4.5 were conducted with temperature set to 0, except for the majority-voting baseline for gpt-4.1, which used temperature 0.7.

\section{\texttt{TextGrad} Implementation Details} \label{appendix:textgrad}
Since TextGrad requires a validation set for prompt optimization, we used 10 samples for training and 10 samples for validation on VLRewardBench.
The reported results on VLRewardBench are based on the remaining 1,227 samples (out of 1,247), as the entire test set was used for other methods.

For Multimodal RewardBench, from the remaining 3,711 samples, we randomly selected 40 examples and split them into 20 for training and 20 for validation.

Following the default configuration used in the official TextGrad examples, we trained for 3 epochs with a batch size of 3.
We then selected the prompts that achieved the highest validation scores for each benchmark and applied the corresponding final prompts uniformly to their test sets.

For TextGrad, the relative inference cost includes both test-time inference and the additional inference required during its training stage.

\section{Prompt Templates} \label{appendix:prompt_template}
Figures~\ref{appendix:vanilla_prompt}–\ref{fig:summarization_prompt} are prompt templates that are used for our experiments.

\section{Function for Extracting Answers}
\begin{lstlisting}[style=pythonstyle, 
caption={A function used to extract judgments from model-generated responses. This version slightly modifies the original Multimodal RewardBench code, adding stricter formatting requirements.}]
def extract_judgment(judgment):
    if "[[A]]" in judgment and "[[B]]" in judgment:
        return "Not judged in the proper format.  [[A,B]]"
    if "[[A]]" in judgment:
        return "A"
    elif "[[B]]" in judgment:
        return "B"
    elif "[A]" in judgment:
        return "A"
    elif "[B]" in judgment:
        return "B"
    else:
        return "Not judged in the proper format."
\end{lstlisting}

\section{Examples from the Vanilla Baseline and \methodselective} \label{appendix:examples}
\begin{figure}[t]
    \centering
    \includegraphics[width=0.5\linewidth]{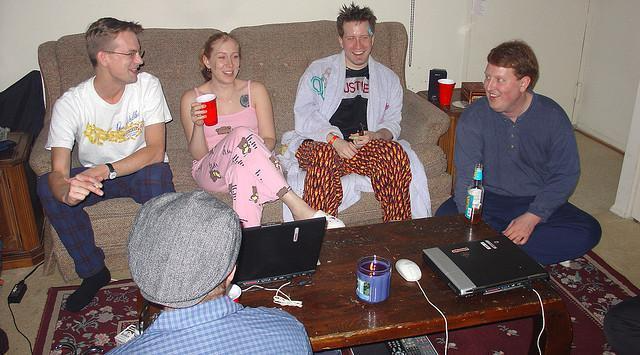}
    \caption{ 
    Image of the test case \texttt{hallucination\_pair-4608} from VLRewardBench.}
    \label{fig:placeholder}
\end{figure}

Figures~\ref{fig:vanilla-eval-prompt}–\ref{appendix:last_prompt} illustrate the actual prompts generated while evaluating a test case \texttt{hallucination\_pair-4608} from VLRewardBench.
The corresponding input image is shown in Figure~\ref{fig:placeholder}.
While the Vanilla baseline incorrectly selects response A, \methodselective correctly identifies response B as the better answer.

\section{Full Meta-Prompts from Figure~\ref{fig:meta_update_example}} \label{appendix:fig7_full_prompts}
Figures~\ref{fig:meta_update_example_fig7_1} and \ref{fig:meta_update_example_fig7_2} present the full meta-prompts corresponding to Figure~\ref{fig:meta_update_example}.
These examples illustrate how the meta-prompt is refined after a single update step, transitioning from an earlier version (Figure~\ref{fig:meta_update_example_fig7_1}) to the updated version (Figure~\ref{fig:meta_update_example_fig7_2}).

\section{Pseudocode of Evaluation Methods}
\label{appendix:pseudocode}
Algorithms~\ref{alg:vanilla_appendix} and ~\ref{alg:sample-specific-appendix} present the pseudocode of the Vanilla and Sample-Specific Prompt baselines.
All procedures (\texttt{Judge} and \texttt{BuildEvalPrompt}) are implemented through LLM prompting.

    \begin{algorithm}[t]
        \caption{\texttt{Vanilla}}
        \label{alg:vanilla_appendix}
        \textbf{Input:} A base LLM $\texttt{LLM}$, a test set $D_{\text{test}}$, and a vanilla prompt $P$ \\
        \textbf{Output:} A set of evaluated results $S$ \\[-0.8em]
        \begin{algorithmic}[1]
            \STATE $S \leftarrow \emptyset$ \hfill $\triangleright$ Initialize a set of evaluation results.
            \FOR{$x \in D_{\text{test}}$ }
                \STATE $y \leftarrow \texttt{Judge}_{\texttt{LLM}}(P, x)$ 
                \STATE $S \leftarrow S \cup \{(x, y)\}$
            \ENDFOR
            \RETURN $S$
        \end{algorithmic}
    \end{algorithm}

    \begin{algorithm}[t]
        \caption{\texttt{Sample-Specific Prompt}}
        \label{alg:sample-specific-appendix}
        \textbf{Input:} A base LLM \texttt{LLM}, a test set $D_{\text{test}}$, and an initial meta-prompt $M_0$ \\
        \textbf{Output:} A set of evaluated results $S$ \\ [-0.8em]
        \begin{algorithmic}[1]
            \STATE $S \leftarrow \emptyset$ \hfill $\triangleright$ Initialize a set of evaluated results.
            \FOR{$x \in D_{\text{test}}$ }
                \STATE $P \leftarrow \texttt{BuildEvalPrompt}_{\texttt{LLM}}(M_0, x)$
                \STATE $y \leftarrow \texttt{Judge}_{\texttt{LLM}}(P, x)$ 
                \STATE $S \leftarrow S \cup \{(x, y)\}$
            \ENDFOR
            \RETURN $S$
        \end{algorithmic}
    \end{algorithm}

\section{Test Case Ordering Results}\label{appendix:ordering_results_all}
Tables~\ref{tab:ordering_results_lwe_vl}–\ref{tab:ordering_results_selective_lwe_mm} show the results of \method and \textit{Selective} \method with gpt-4.1 on each benchmark with three random orders of test cases.
Tables~\ref{tab:ordering_results_selective_lwe_vl_gemini}–\ref{tab:ordering_results_selective_lwe_mm_claude} provide the corresponding \methodselective results using gemini-2.5-pro and claude-sonnet-4.5 on each benchmark.

\section{\method on Inconsistent \vs Consistent Subsets}\label{appendix:inconsistent_consistent_ablation}
We investigate why \method achieves higher accuracy than \methodselective on VLRewardBench.
An inspection of the inconsistent and consistent subsets shows that VLRewardBench contains a relatively large number of vanilla-consistent but incorrect examples (999 - 659 = 340 examples) compared to that of MMRewardBench (863 - 746 = 117 examples) (Table~\ref{tab:selection_analysis_consistent}).
We hypothesize that updates on these cases provide additional supervisory signal for the meta-prompt, which may contribute to the higher accuracy observed for LWE.

However, this effect appears to impact accuracy only, not pair accuracy.
Across both benchmarks, \methodselective consistently achieves higher pair accuracy, which reliably measures the evaluator’s ability and robustness to position bias.
Moreover, \method exhibits substantially larger improvements on the vanilla-inconsistent subsets (Table~\ref{tab:selection_analysis}) compared to the consistent subsets (Table~\ref{tab:selection_analysis_consistent}).
Taken together, these patterns suggest that, despite the presence of consistent but wrong cases, the vanilla inconsistency signal remains a useful and reliable criterion for determining when updates should be applied.

\begin{table}[t]
\centering
\resizebox{\columnwidth}{!}{
\begin{tabular}{lcc}
\hline
\textbf{Method} & \textbf{VLRewardBench} & \textbf{MMRewardBench} \\
\hline
Vanilla & 0.504 (125/248) & 0.453 (62/137) \\
\textit{Selective} LWE & 0.762 (189/248) & 0.672 (92/137) \\ 
\hline
\end{tabular}}
\caption{\textbf{Accuracy on the inconsistent subsets.} \textit{Selective} LWE leads to more accurate judgments.
Results correspond to a single run (run~0) from Tables~\ref{tab:ordering_results_selective_lwe_vl} and ~\ref{tab:ordering_results_selective_lwe_mm}.}
\label{tab:selection_analysis}
\end{table}

\begin{table}[t]
\centering
\resizebox{\columnwidth}{!}{
\begin{tabular}{lcc}
\hline
\textbf{Method} & \textbf{VLRewardBench} & \textbf{MMRewardBench} \\
\hline
Vanilla & 0.660 (659/999) & 0.864 (746/863) \\
LWE & 0.734 (733/999) & 0.846 (730/863) \\ 
\hline
\end{tabular}}
\caption{\textbf{Accuracy on the consistent subsets.} As an ablation, we apply LWE to consistent examples; the resulting performance gains are modest compared to those observed on inconsistent cases.}
\label{tab:selection_analysis_consistent}
\end{table}

\section{Comparison with Previous Works} \label{appendix:previous_works}
Table~\ref{tab:comparison} summarizes the comparison with previous studies.

\section{Data Licensing}
We use publicly available datasets under research-only and permissive licenses.
VLRewardBench is released for research use only.
Multimodal RewardBench is licensed under the Creative Commons Attribution-NonCommercial (CC BY-NC)\footnote{\url{https://creativecommons.org/licenses/by-nc/4.0/}},
Copyright (c) Meta Platforms, Inc. and affiliates.
TextGrad and Dynamic Cheatsheet are released under the MIT License\footnote{\url{https://opensource.org/licenses/mit}}.
All datasets and codebases were used solely for non-commercial, academic research in compliance with their respective licenses.

\clearpage

\begin{table}[t]
\vspace{2em}
\centering
\resizebox{\columnwidth}{!}{
    \begin{tabular}{lccccc}
    \hline
    \textbf{LWE} & \textbf{run0} & \textbf{run1} & \textbf{run2} & \textbf{mean} & \textbf{std.} \\
    \hline
    Acc. & 0.746 & 0.713 & 0.776 & \textbf{0.745} &\textbf{ 0.031} \\
    Cons. & 0.808 & 0.789 & 0.818 &\textbf{ 0.805} & \textbf{0.015} \\
    PairAcc. & 0.650 & 0.610 & 0.677 &\textbf{ 0.646} &\textbf{ 0.033} \\
    \hline
    \end{tabular}
}
\caption{Performance of \method\ under different orderings of VLRewardBench. The table reports the mean and standard deviation across three independent runs.}
\label{tab:ordering_results_lwe_vl}
\end{table}

\begin{table}[t]
\vspace{4.5em}
\centering
\resizebox{\columnwidth}{!}{
    \begin{tabular}{lccccc}
    \hline
    \textbf{LWE} & \textbf{run0} & \textbf{run1} & \textbf{run2} & \textbf{mean} & \textbf{std.} \\
    \hline
    Acc. & 0.803 & 0.806 & 0.789 & \textbf{0.799} & \textbf{0.009} \\
    Cons. & 0.839 & 0.850 & 0.850 & \textbf{0.846} & \textbf{0.006} \\
    PairAcc. & 0.727 & 0.727 & 0.727 & \textbf{0.727} & \textbf{0.000} \\
    \hline
    \end{tabular}
}
\caption{Performance of \method\ under different orderings of MMRewardBench. The table reports the mean and standard deviation across three independent runs.}
\label{tab:ordering_results_lwe_mm}
\end{table}

\begin{table}[t]
\vspace{4.5em}
\centering
\resizebox{\columnwidth}{!}{
\begin{tabular}{lccccc}
\hline
\textbf{\textit{Selective} LWE} & \textbf{run0} & \textbf{run1} & \textbf{run2} & \textbf{mean} & \textbf{std.} \\
\hline
Acc. & 0.680 & 0.674 & 0.674 & \textbf{0.676 }& \textbf{0.003} \\
Cons. & 0.945 & 0.938 & 0.937 & \textbf{0.940} & \textbf{0.004} \\
PairAcc. & 0.653 & 0.649 & 0.642 &\textbf{ 0.648 }& \textbf{0.006} \\
\hline
\end{tabular}
}
\caption{Performance of \textit{Selective} \method\ under different orderings of VLRewardBench. The table reports the mean and standard deviation across three independent runs.}
\label{tab:ordering_results_selective_lwe_vl}
\end{table}

\begin{table}[t]
\vspace{5em}
\centering
\resizebox{\columnwidth}{!}{
\begin{tabular}{lccccc}
\hline
\textbf{\textit{Selective} LWE} & \textbf{run0} & \textbf{run1} & \textbf{run2} & \textbf{mean} & \textbf{std.} \\
\hline
Acc. & 0.838 & 0.837 & 0.832 & \textbf{0.836} & \textbf{0.003} \\
Cons. & 0.940 & 0.954  & 0.946 &\textbf{ 0.947} & \textbf{0.007 }\\
PairAcc. & 0.806 & 0.813 & 0.805 &\textbf{ 0.808}&\textbf{ 0.004} \\

\hline
\end{tabular}
}
\caption{Performance of \textit{Selective} \method\ under different orderings of MMRewardBench. The table reports the mean and standard deviation across three independent runs.}
\label{tab:ordering_results_selective_lwe_mm}
\end{table}

\begin{table}
\centering
\resizebox{\columnwidth}{!}{
\begin{tabular}{lccccc}
\hline
\textbf{\textit{Selective} LWE} & \textbf{run0} & \textbf{run1} & \textbf{run2} & \textbf{mean} & \textbf{std.} \\
\hline
Acc. & 0.770	& 0.767	& 0.767			& \textbf{0.768}	& \textbf{0.001} \\
Cons. & 0.955	& 0.964	& 0.946			& \textbf{0.955}	& \textbf{0.009} \\
PairAcc. & 0.746	& 0.748	& 0.739			& \textbf{0.744}	& \textbf{0.005} \\
\hline
\end{tabular}
}
\caption{Performance of \methodselective under different orderings of VLRewardBench with gemini-2.5-pro. The table reports the mean and standard deviation across three independent runs.}
\label{tab:ordering_results_selective_lwe_vl_gemini}
\end{table}

\begin{table}
\centering
\resizebox{\columnwidth}{!}{
\begin{tabular}{lccccc}
\hline
\textbf{\textit{Selective} LWE} & \textbf{run0} & \textbf{run1} & \textbf{run2} & \textbf{mean} & \textbf{std.} \\
\hline
Acc. &0.865	    &0.862	&0.868	&\textbf{0.865}	&\textbf{0.003} \\
Cons. &0.975	&0.959	&0.972	&\textbf{0.969}	&\textbf{0.009} \\
PairAcc. &0.857	&0.846	&0.853	&\textbf{0.852}	&\textbf{0.006} \\
\hline
\end{tabular}
}
\caption{Performance of \methodselective under different orderings of MMRewardBench with gemini-2.5-pro. The table reports the mean and standard deviation across three independent runs.}
\label{tab:ordering_results_selective_lwe_mm_gemini}
\end{table}

\begin{table}
\centering
\resizebox{\columnwidth}{!}{
\begin{tabular}{lccccc}
\hline
\textbf{\textit{Selective} LWE} & \textbf{run0} & \textbf{run1} & \textbf{run2} & \textbf{mean} & \textbf{std.} \\
\hline
Acc. &0.705	&0.704	&0.701	&\textbf{0.703}	&\textbf{0.002} \\
Cons. &0.878	&0.885	&0.881	&\textbf{0.881}	&\textbf{0.003} \\
PairAcc. &0.646	&0.647	&0.650	&\textbf{0.648}	&\textbf{0.002} \\
\hline
\end{tabular}
}
\caption{Performance of \methodselective under different orderings of VLRewardBench with claude-sonnet-4.5. The table reports the mean and standard deviation across three independent runs.}
\label{tab:ordering_results_selective_lwe_vl_claude}
\end{table}

\begin{table}
\centering
\resizebox{\columnwidth}{!}{
\begin{tabular}{lccccc}
\hline
\textbf{\textit{Selective} LWE} & \textbf{run0} & \textbf{run1} & \textbf{run2} & \textbf{mean} & \textbf{std.} \\
\hline
Acc. &0.841	&0.837	&0.791	&\textbf{0.823}	&\textbf{0.028} \\
Cons. &0.912	&0.893	&0.831	&\textbf{0.879}	&\textbf{0.042} \\
PairAcc. &0.795	&0.782	&0.736	&\textbf{0.771}	&\textbf{0.031} \\
\hline
\end{tabular}
}
\caption{Performance of \methodselective under different orderings of MMRewardBench with claude-sonnet-4.5. The table reports the mean and standard deviation across three independent runs.}
\label{tab:ordering_results_selective_lwe_mm_claude}
\end{table}

\clearpage
\begin{table*}[t]
\centering
\resizebox{\textwidth}{!}{
\begin{tabular}{lcccc}
\hline
\textbf{Method} & \textbf{Updates Text} & \textbf{Selectively Updates Text} & \textbf{Tests Sequentially} &  \textbf{Explicitly Generates Sample-specific Prompts}  \\ 
\hline
TextGrad & on a \textit{validation} set & \xmark & \xmark & \xmark \\
Dynamic Cheatsheet & on a \textit{test} set & \xmark & \cmark & \xmark \\
\textit{Selective} LWE & on a \textit{test} set & \cmark & partially \cmark & \cmark \\
\hline
\end{tabular}}
\caption{Comparison with previous works.}
\label{tab:comparison}
\end{table*}

\newpage
\newcommand{\promptfont}{\ttfamily\small}

\begin{figure*}[t]
\centering
\begin{tcolorbox}[width=\textwidth,
  enhanced, breakable,
  colback=blue!2, colframe=blue!60!black,
  boxrule=0.8pt, sharp corners,
  fontupper=\ttfamily\small
]
\promptfont
Please act as an impartial judge and evaluate the quality of the responses provided by two AI assistants to the user question displayed below. You should choose the assistant that follows the user's instructions and answers the user's question better. Your evaluation should consider factors such as the helpfulness, relevance, accuracy, depth, creativity, and level of detail of their responses. Begin your evaluation by comparing the two responses and provide a short explanation. Avoid any position biases and ensure that the order in which the responses were presented does not influence your decision. Do not allow the length of the responses to influence your evaluation. Do not favor certain names of the assistants. Be as objective as possible. After providing your explanation, output your final verdict by strictly following this format: "[[A]]" if assistant A is better, "[[B]]" if assistant B is better. \\

[User Question]\\
\{question\} \\

[The Start of Assistant A's Answer] \\
\{answer\_a\} \\
\texttt{[The End of Assistant A's Answer]} \\

[The Start of Assistant B's Answer] \\
\{answer\_b\} \\
\texttt{[The End of Assistant B's Answer]}
\end{tcolorbox}
\caption{\texttt{Vanilla} evaluation prompt. We follow the prompt of \cite{yasunaga2025multimodalrewardbenchholisticevaluation}.}
\label{appendix:vanilla_prompt}
\end{figure*}

\begin{figure*}[t]
\centering
\begin{tcolorbox}[width=\textwidth,
  enhanced, breakable,
  colback=blue!2, colframe=blue!60!black,
  boxrule=0.8pt, sharp corners,
  fontupper=\ttfamily\small
]
\promptfont
Please act as an impartial judge and evaluate the quality of the responses provided by two AI assistants to the user question displayed below. You should choose the assistant that follows the user's instructions and answers the user's question better. Your evaluation should consider factors such as the helpfulness, relevance, accuracy, depth, creativity, and level of detail of their responses. Begin your evaluation by comparing the two responses and provide a short explanation. Avoid any position biases and ensure that the order in which the responses were presented does not influence your decision. Do not allow the length of the responses to influence your evaluation. Do not favor certain names of the assistants. Be as objective as possible. After providing your explanation, output your final verdict by strictly following this format: "[[A]]" if assistant A is better, "[[B]]" if assistant B is better. \\

[User Question]\\
\{question\} \\

[The Start of Assistant A's Answer] \\
\{answer\_a\} \\
\texttt{[The End of Assistant A's Answer]} \\

[The Start of Assistant B's Answer] \\
\{answer\_b\} \\
\texttt{[The End of Assistant B's Answer]} \\

Please reason in a step-by-step manner before giving a response. (You now have an opportunity to reason privately; your next response will not be evaluated.)
\end{tcolorbox}
\caption{\texttt{Chain-of-Thought (CoT)} evaluation prompt. We adopt the Chain-of-Thought (CoT) evaluation prompt from the OpenAI Evals repository~\citep{openai-evals}.}
\label{appendix:cot_prompt}
\end{figure*}

\begin{figure*}[t]
\centering
\begin{tcolorbox}[width=\textwidth,
  enhanced, breakable,
  colback=blue!2, colframe=blue!60!black,
  boxrule=0.8pt, sharp corners,
]
\promptfont
Generate a prompt for an evaluator that includes example-specific evaluation criteria and step-by-step evaluation procedures. The evaluator will base their evaluation on the prompt and it will help the evaluator accurately judge the correctness and reasoning quality of given responses, and identify which answer is better. \\

Generated evaluation prompt MUST include \\
- evaluation criteria specific for the given example \\
- evaluation steps specific for the given example to induce an accurate and reliable judgment. \\

Additionally, the evaluation prompt MUST instruct the evaluator that their final judgment must be expressed only in one of the following two formats: \\
`[[A]]' or `[[B]]'. \\

Output exactly ONLY THE PROMPT. \\
\end{tcolorbox}
\caption{Initial meta prompt. We use this prompt as the static meta prompt in \texttt{Sample-Specific Prompt} and the initial meta prompt in LWE and \textit{Selective} LWE (\texttt{BuildEvalPrompt}).}
\label{appendix:initial_meta_prompt}
\end{figure*}

\clearpage

\begin{figure*}[t]
\centering
\begin{tcolorbox}[width=\textwidth,
  enhanced, breakable,
  colback=blue!2, colframe=blue!60!black,
  boxrule=0.8pt, sharp corners,
  fontupper=\ttfamily\small
]
\promptfont
You will be given: \\
An evaluation prompt — the instructions or criteria for judging \\
A given example — a case where you have already made a judgment based on that prompt \\
 \\
Your task: \\
Evaluate the correctness and reasoning quality of the given judgment. \\
Based on the inspection, update your [[Learned tips for future prompt optimization]] based on the current "meta\_prompt". \\
 \\
Your output must have three parts: \\
\texttt{[[[Score (1–5)]]]} \\
5 = Judgment is entirely correct, thorough, and follows the evaluation prompt exactly \\
4 = Mostly correct, with only minor reasoning gaps \\
3 = Partially correct, but with noticeable flaws or missing justification \\
2 = Largely incorrect, due to poor reasoning or serious omissions \\
1 = Fundamentally wrong, fails to follow the evaluation prompt \\
 \\
\texttt{[[[Binary label]]]} \\
"Absolutely confident the judgment is correct" — if reasoning is strong, evidence-based, and follows the criteria without major gaps \\
"Not sure" — if there are reasoning flaws, possible alternative interpretations, or insufficient evidence \\
 \\
\texttt{[[[Learned tips for future prompt optimization]]]} \\
After reading the given example and its judgment, identify specific points to improve in the reasoning, coverage, or clarity. \\
Reflect on these findings to propose concrete adjustments to future evaluation prompts (e.g., clarifying ambiguous criteria, adding explicit checks for evidence, requiring certain logical structures) \\
Capture any patterns that might cause repeated errors or inconsistencies so they can be addressed in later prompts. \\
You might refer to the current meta prompt and identify additional essential tips that are not addressed in the current meta prompt. \\
 \\
**Evaluation Steps (You must follow these before scoring)**: \\
Step 1: Restate the evaluation criteria in your own words to ensure you understand them. \\
Step 2: Read the given example judgment carefully and identify its main claim(s) and supporting reasoning. \\
Step 3: Compare the judgment's reasoning with the evaluation criteria; look for matches, missing points, contradictions, or misinterpretations. \\
Step 4: Inspect for logical fallacies, including: \\
    - Unsupported generalizations \\
    - False cause (confusing correlation with causation) \\
    - Strawman misrepresentations of the criteria \\
    - Circular reasoning \\
    - Irrelevant evidence \\
Step 5: Pinpoint exact areas for improvement in the judgment's reasoning or evidence use. \\
Step 6: Assign a score based on the completeness and accuracy of the reasoning relative to the prompt. \\
Step 7: Decide on the binary confidence label. \\
Step 8: Reflect on how these identified weaknesses could be prevented through clearer or more detailed future evaluation prompts, and write down 1–3 learned tips. \\
 \\
\texttt{[[[The Start of Current Meta Prompt]]]} \\
\{meta\_prompt\} \\
\texttt{[[[The End of Current Meta Prompt]]]} \\
 \\
\texttt{[[[The Start of Evaluation Prompt \& Judgment]]]} \\
\{evaluation\_prompt\} \\
\\
\texttt{[Start of Judgment]} \\
\{judgment\} \\
\texttt{[The End of Judgment]} \\
\texttt{[[[The End of Evaluation Prompt \& Judgment]]]} \\
 \\
Strictly follow the **Output format**: \\
\{\{"score": score, "label": label, "learned tips": tips, "reasoning": reasoning\_or\_explanation\}\}
\end{tcolorbox}
\caption{Prompt for feedback generation (\texttt{Feedback}).}
\label{feedback_generation}
\end{figure*}

\clearpage

\begin{figure*}[t]
\centering
\begin{tcolorbox}[width=\textwidth,
  enhanced, breakable,
  colback=blue!2, colframe=blue!60!black,
  boxrule=0.8pt, sharp corners,
  fontupper=\ttfamily\small
]
\promptfont
Optimize your current meta prompt. You need to refer to feedback. You need to improve the meta prompt. The goal of meta prompt is to generate an evaluation prompt that helps more accurate and reliable judgments. A good evaluation prompt needs better example-specific evaluation rubrics and evaluation steps. Make sure not to repeat any tips that are already provided. Instead, write only additional tips that are general, reusable, and useful for evaluation. Focus on refining and improving the quality of your learned tips and experiences. \\
 \\
\texttt{[[[The Start of Current Meta Prompt]]]} \\
\{meta\_prompt\} \\
\texttt{[[[The End of Current Meta Prompt]]]} \\
 \\
\texttt{[[[The Start of Batch examples \& Meta Feedback]]]} \\
\{batch\} \\
\texttt{[[[The End of Batch examples \& Meta Feedback]]]} \\
 \\
Again, optimize your current meta prompt based on the feedback. \\
\texttt{[[[Optimized Meta Prompt]]]} \\

\end{tcolorbox}
\caption{Prompt template for refining a meta prompt (\texttt{RefineMetaPrompt}). Each “batch” includes the current evaluation prompts, examples, judgments, and feedback.}
\label{appendix:refinemetaprompt}
\end{figure*}

\begin{figure*}[t]
\centering
\begin{tcolorbox}[width=\textwidth,
  enhanced, breakable,
  colback=blue!2, colframe=blue!60!black,
  boxrule=0.8pt, sharp corners,
  fontupper=\ttfamily\small
]
\promptfont
\texttt{[User Question]}\\
\{question\}\\
\\
\texttt{[The Start of Assistant A's Answer]}
\\
\{answer\_a\}\\
\texttt{[The End of Assistant A's Answer]}\\
\\
\texttt{[The Start of Assistant B's Answer]}\\
\{answer\_b\}\\
\texttt{[The End of Assistant B's Answer]}
\end{tcolorbox}
\caption{Example placeholder. The above placeholder with an example is appended at the end of each evaluation prompt to ensure stable generation.}
\label{appendix:example_placeholder}
\end{figure*}

\begin{figure*}[t]
\centering
\begin{tcolorbox}[width=\textwidth,
  enhanced, breakable,
  colback=blue!2, colframe=blue!60!black,
  boxrule=0.8pt, sharp corners,
  fontupper=\ttfamily\small
]
\promptfont
You are a helpful assistant that summarizes and shortens text. Your current meta prompt is too long. \\
\texttt{[[[The Start of Current Meta Prompt]]]} \\
\{meta\_prompt\} \\
\texttt{[[[The End of Current Meta Prompt]]]} \\
 \\
Condense the meta prompt so it is roughly HALF as long, but still structured and keeping the key details. \\
ONLY output the shortened meta prompt. \\
\end{tcolorbox}
\caption{Summarization prompt. As described in Appendix~\ref{subsection:details}, when a meta prompt exceeds a predefined length, it is summarized by the model itself.} \label{fig:summarization_prompt}
\end{figure*}

\clearpage

\begin{figure*}[t]
\centering
\begin{tcolorbox}[colback=gray!5, colframe=gray!50, 
                  boxrule=0.5pt, arc=2mm, left=4pt, right=4pt, 
                  top=3pt, bottom=3pt, title=Example with the vanilla evaluation prompt,
                  width=\textwidth]
\noindent Please act as an impartial judge and evaluate the quality of the responses provided by two AI assistants to the user question displayed below. You should choose the assistant that follows the user's instructions and answers the user's question better. Your evaluation should consider factors such as the helpfulness, relevance, accuracy, depth, creativity, and level of detail of their responses. Begin your evaluation by comparing the two responses and provide a short explanation. Avoid any position biases and ensure that the order in which the responses were presented does not influence your decision. Do not allow the length of the responses to influence your evaluation. Do not favor certain names of the assistants. Be as objective as possible. After providing your explanation, output your final verdict by strictly following this format: \texttt{"[[A]]"} if assistant A is better, \texttt{"[[B]]"} if assistant B is better.\\
\\
\texttt{[User Question]}\\
Describe this image in detail.\\
\\
\texttt{[The Start of Assistant A's Answer]}\\
The image shows a group of friends sitting on a couch in a living room, enjoying each other's company. There are five people in total, including a pretty young lady and four men. They appear to be dressed in casual or pajama attire, suggesting a relaxed atmosphere. Snacks or food can be seen on the table, adding to the cozy vibe. As they share a funny story, some drinks can also be observed in the scene, with three cups and a bottle placed around the room, likely adding to the merriment. On a table, there are two laptops and a mouse, hinting that the group may be engaged in some kind of electronic activity or entertainment as well.\\
\texttt{[The End of Assistant A's Answer]}\\
\\
\texttt{[The Start of Assistant B's Answer]}\\
The image shows a group of friends sitting on a couch in a living room, enjoying each other's company. There are five people in total, including a pretty young lady and four men. They appear to be dressed in casual or pajama attire, suggesting a relaxed atmosphere.\\
\\
As they share a funny story, some drinks can be observed in the scene, with three cups and a bottle placed around the room, likely adding to the merriment. On a table, there are two laptops and a mouse, hinting that the group may be engaged in some kind of electronic activity or entertainment as well.\\
\texttt{[The End of Assistant B's Answer]}
\end{tcolorbox}
\caption{Example with the vanilla evaluation prompt used for pairwise judgment.}
\label{fig:vanilla-eval-prompt}
\end{figure*}

\begin{figure*}[t]
\centering
\begin{tcolorbox}[colback=gray!5, colframe=gray!50, 
                  boxrule=0.5pt, arc=2mm, left=4pt, right=4pt, 
                  top=3pt, bottom=3pt, title=Example of a judgment with the vanilla evaluation prompt,
                  width=\textwidth]
\noindent Assistant A and Assistant B provide very similar descriptions of the image, both noting the group of five friends, the casual or pajama attire, the presence of drinks, laptops, and the relaxed, social atmosphere. However, Assistant A's response is slightly more detailed, mentioning the snacks or food on the table and describing the overall vibe as ``cozy.'' This adds a bit more depth and context to the scene, making the description more complete and helpful. Both responses are accurate and relevant, but Assistant A's answer offers a touch more detail and completeness.\\
\\
\texttt{[[A]]}
\end{tcolorbox}
\caption{Example of a vanilla judgment by the prompt in Figure~\ref{fig:vanilla-eval-prompt}.}
\label{fig:vanilla-judgment}
\end{figure*}

\begin{figure*}[t]
\centering
\begin{tcolorbox}[breakable, width=0.95\textwidth, colback=blue!5, colframe=blue!50, boxrule=0.5pt, arc=2mm, title=Example of a meta prompt]
[[[Optimized Meta Prompt]]]\\
\\
**Meta Prompt for Generating High-Quality Evaluation Prompts**\\
\\
Generate an evaluation prompt for an evaluator that produces more accurate and reliable judgments by incorporating example-specific rubrics and detailed evaluation steps. The evaluation prompt you generate MUST:\\
\\
- Include evaluation criteria tailored to the specific example, focusing on observable evidence, internal consistency, and avoidance of unsupported speculation or irrelevant details.\\
- Provide step-by-step evaluation procedures that guide the evaluator to systematically compare each answer against the example, cross-checking all claims with the available evidence and justifying their choice with reference to specific elements.\\
- Instruct the evaluator to penalize both the omission of significant details and the inclusion of non-existent or speculative information.\\
- Emphasize the importance of directness, clarity, and relevance, and require explicit reference to the evidence supporting their decision.\\
- Require that the evaluator’s final judgment be expressed only in one of the following two formats: `[[A]]' or `[[B]]'.\\
\\
**In addition, based on learned experience and feedback, incorporate these further general, reusable tips to improve evaluation quality. Do NOT repeat any tips already provided above; instead, add only new, generalizable, and actionable tips that further refine evaluation quality:**\\
\end{tcolorbox}
\caption{Example of a meta prompt (Part 1).}
\end{figure*}

\begin{figure*}[t]\ContinuedFloat
\centering
\begin{tcolorbox}[breakable, width=0.95\textwidth, colback=blue!5, colframe=blue!50, boxrule=0.5pt, arc=2mm]
**Additional Tips for Further Improvement:**\\
\\
- Require evaluators to **explicitly restate the observable evidence at the start of their evaluation** to anchor all subsequent judgments in the provided data.\\
- Instruct evaluators to **explicitly check for and penalize misinterpretation of labels, diagrams, or visual cues** (e.g., confusing radius and diameter, or misreading image features), as these are common sources of error.\\
- Require evaluators to **explicitly verify that the selected answer is at least partially relevant to the actual example and prompt**, not just internally consistent or well-reasoned.\\
- Instruct evaluators to **call out and penalize even minor speculative or fabricated details in both answers**, not just omissions, and to reward concise answers that avoid unsupported elaborations.\\
- Add a step for evaluators to **explicitly check for off-topic or irrelevant answers**; if both answers are off-topic, reward the one that at least attempts to address the prompt or acknowledge when neither is correct.\\
- Require evaluators to **briefly justify why the rejected answer is incorrect or less preferred**, especially when visual or factual evidence is available.\\
- Encourage evaluators to **distinguish between errors due to plausible but unsupported assumptions and outright fabrications**, and to penalize accordingly.\\
- Instruct evaluators to **reward transparent, evidence-limited reasoning**—for example, when an answer acknowledges the limits of the available evidence or the possibility that neither option is fully correct.\\
- Remind evaluators to **explicitly reference the evidence in their reasoning, not just in the final answer**, and to ensure that all preferences and penalizations are justified by direct reference to the example.\\
\\
**Meta-Reminders:**\\
- Do not judge the example contents themselves.\\
- Do not include or reference the example contents in your evaluation prompt.\\
- Follow these steps and criteria strictly for every evaluation.\\
\\
**Goal:**\\
The resulting evaluation prompt should enable evaluators to make more accurate, reliable, and evidence-grounded judgments, with explicit, example-specific rubrics and procedures, and with all preferences and penalizations justified by direct reference to the evidence.
\end{tcolorbox}
\caption{Example of a meta prompt (Part 2).}
\label{fig:meta-prompt}
\end{figure*}

\begin{figure*}[t]
\centering
\begin{tcolorbox}[breakable, width=0.95\textwidth, colback=blue!5, colframe=blue!50, boxrule=0.5pt, arc=2mm, title=Example of a sample-specific evaluation prompt generated by a meta prompt]
"**Evaluation Prompt Template**\\
\\
**Step 1: Restate the Observable Evidence**\\
Begin your evaluation by explicitly restating the observable evidence from the example. List the key visual or factual elements that are directly present and verifiable.\\
\\
**Step 2: Apply Example-Specific Evaluation Criteria**\\
Evaluate each answer using the following criteria, tailored to the example:\\
- **Accuracy and Completeness:** Does the answer accurately describe all significant observable details? Are any important elements omitted or misrepresented?\\
- **Internal Consistency:** Is the answer logically consistent within itself, without contradictions?\\
- **Relevance:** Does the answer focus only on details that are directly observable and relevant to the example, avoiding speculation or irrelevant information?\\
- **Evidence-Based Reasoning:** Are all claims in the answer directly supported by the observable evidence? Is there any unsupported speculation or fabrication?\\
- **Clarity and Directness:** Is the answer clear, concise, and free from unnecessary elaboration or ambiguity?\\
\\
**Step 3: Systematic Comparison and Cross-Checking**\\
For each answer:\\
- Systematically compare every claim or detail against the observable evidence.\\
- Explicitly check for and penalize any misinterpretation of visual cues, labels, or features (e.g., miscounting objects, misidentifying items).\\
- Penalize both the omission of significant details and the inclusion of non-existent or speculative information.\\
- Explicitly verify that the answer is at least partially relevant to the actual example and prompt, not just internally consistent.\\
- Call out and penalize even minor speculative or fabricated details, and reward concise answers that avoid unsupported elaborations.\\
- Check for off-topic or irrelevant content; if both answers are off-topic, reward the one that at least attempts to address the prompt or acknowledge when neither is correct.\\
\\
**Step 4: Justification and Preference**\\
- For the answer you prefer, justify your choice by referencing specific elements of the observable evidence and explaining how the answer meets the criteria above.\\
- Briefly justify why the rejected answer is incorrect or less preferred, especially when visual or factual evidence is available.\\
- Distinguish between errors due to plausible but unsupported assumptions and outright fabrications, and penalize accordingly.\\
- Reward transparent, evidence-limited reasoning (e.g., when an answer acknowledges the limits of the available evidence or the possibility that neither option is fully correct).\\
- Ensure that all preferences and penalizations are justified by direct reference to the evidence, not just in the final answer but throughout your reasoning.\\
\\
\end{tcolorbox}
\caption{Example of a sample-specific evaluation prompt generated by the meta prompt in Figure~\ref{fig:meta-prompt} (Part 1).}
\label{fig:eval_prompt_fig16}
\end{figure*}

\begin{figure*}[t]\ContinuedFloat
\centering
\begin{tcolorbox}[breakable, width=0.95\textwidth, colback=blue!5, colframe=blue!50, boxrule=0.5pt, arc=2mm]
**Step 5: Final Judgment**\\
Express your final judgment in one of the following two formats only:\\
- \texttt{[[A]]}\\
- \texttt{[[B]]}\\

**Meta-Reminders:**\\
- Do not judge the example contents themselves.\\
- Do not include or reference the example contents in your evaluation prompt.\\
- Follow these steps and criteria strictly for every evaluation.\\
\\
\texttt{[User Question]}\\
Describe this image in detail.\\
\\
\texttt{[The Start of Assistant A's Answer]}\\
The image shows a group of friends sitting on a couch in a living room, enjoying each other's company. There are five people in total, including a pretty young lady and four men. They appear to be dressed in casual or pajama attire, suggesting a relaxed atmosphere. Snacks or food can be seen on the table, adding to the cozy vibe. As they share a funny story, some drinks can also be observed in the scene, with three cups and a bottle placed around the room, likely adding to the merriment. On a table, there are two laptops and a mouse, hinting that the group may be engaged in some kind of electronic activity or entertainment as well.\\
\texttt{[The End of Assistant A's Answer]}\\
\\
\texttt{[The Start of Assistant B's Answer]}\\
The image shows a group of friends sitting on a couch in a living room, enjoying each other's company. There are five people in total, including a pretty young lady and four men. They appear to be dressed in casual or pajama attire, suggesting a relaxed atmosphere.\\
\\
As they share a funny story, some drinks can be observed in the scene, with three cups and a bottle placed around the room, likely adding to the merriment. On a table, there are two laptops and a mouse, hinting that the group may be engaged in some kind of electronic activity or entertainment as well.\\
\texttt{[The End of Assistant B's Answer]}
\end{tcolorbox}
\caption{Example of a sample-specific evaluation prompt generated by the meta prompt in Figure~\ref{fig:meta-prompt} (Part 2).}
\end{figure*}

\begin{figure*}[t]
\centering
\begin{tcolorbox}[colback=blue!5, colframe=blue!50, 
                  boxrule=0.5pt, arc=2mm, left=4pt, right=4pt, 
                  top=3pt, bottom=3pt, title=Example of a judgment generated by a sample-specific evaluation prompt,
                  width=\textwidth]
\noindent 

**Step 1: Restate the Observable Evidence**\\
- Five people are present: four sitting on a couch (three men, one woman) and one person (gender unclear from this angle) sitting on the floor with their back to the camera.\\
- The people are dressed in casual or pajama-like attire.\\
- The setting is a living room with a couch, a wooden coffee table, and a rug.\\
- On the coffee table: two open laptops, a computer mouse, a candle, some cables, and a drink bottle.\\
- Several people are holding red cups; one person on the floor has a blue can or bottle.\\
- The group appears to be smiling and engaged in conversation or laughter.\\
- The room has a relaxed, informal atmosphere.\\
\\
**Step 2: Apply Example-Specific Evaluation Criteria**\\
- **Accuracy and Completeness:** Both answers mention five people, casual/pajama attire, a living room, laptops, a mouse, drinks, and a relaxed atmosphere. Assistant A uniquely mentions snacks/food on the table, which is not clearly visible in the image. Both mention a "pretty young lady," which is subjective and not strictly observable.\\
- **Internal Consistency:** Both answers are internally consistent.\\
- **Relevance:** Both focus on observable details, but both include some speculation (e.g., "likely adding to the merriment," "may be engaged in some kind of electronic activity").\\
- **Evidence-Based Reasoning:** Both answers are mostly evidence-based, but A includes a fabricated detail about snacks/food, which is not supported by the image.\\
- **Clarity and Directness:** Both are clear and concise, but A is slightly more verbose due to the extra (unsupported) detail.\\
\\
**Step 3: Systematic Comparison and Cross-Checking**\\
- Both answers correctly identify the number of people, the general attire, the presence of laptops, and the drinks.\\
- Both mention a "pretty young lady," which is subjective and not strictly evidence-based.\\
- Both mention drinks, cups, and a bottle, which are visible.\\
- Both mention the possibility of electronic activity, which is a reasonable inference from the laptops.\\
- Assistant A adds "snacks or food can be seen on the table," which is not supported by the image—there is a candle, but no visible food or snacks.\\
- Both answers are relevant to the prompt, but A includes a minor fabrication.\\
\\
**Step 4: Justification and Preference**\\
- **Preferred Answer:** Assistant B is preferred because it avoids the unsupported claim about snacks/food, sticking more closely to the observable evidence. While both answers include some minor speculation and subjective language, B does not fabricate any details about the scene.\\
- **Rejected Answer:** Assistant A is less preferred due to the inclusion of a fabricated detail (snacks/food) that is not visible in the image. This is a minor but clear deviation from evidence-based reasoning.\\
\\
**Step 5: Final Judgment**\\
\texttt{[[B]]}
\end{tcolorbox}
\caption{Example of a judgment generated by the sample-specific evaluation prompt in Figure~\ref{fig:eval_prompt_fig16}.}
\label{appendix:last_prompt}
\end{figure*}

\begin{figure*}[t]
\centering
\begin{tcolorbox}[breakable, width=0.95\textwidth, colback=blue!5, colframe=blue!50, boxrule=0.5pt, arc=2mm, title=Example of a meta prompt before an update]
[[[Optimized Meta Prompt]]]\\
\\
Generate an evaluation prompt for an evaluator that includes example-specific evaluation criteria and step-by-step evaluation procedures. The evaluator will use this prompt to accurately judge the correctness and reasoning quality of given responses, and to identify which answer is better.\\
\\
The evaluation prompt you generate MUST include:\\
- Evaluation criteria tailored to the specific example, covering all aspects required for a thorough and precise judgment.\\
- Step-by-step evaluation procedures that are specific to the example and guide the evaluator to a reliable and accurate decision.\\
\\
Additionally, the evaluation prompt MUST instruct the evaluator that their final judgment must be expressed only in one of the following two formats:\\
`[[A]]' or `[[B]]'.\\
\\
When generating the evaluation prompt, incorporate the following additional tips to further improve accuracy and reliability (do NOT repeat tips already provided in the current meta prompt or previous feedback):\\
\\
**Additional Tips:**\\
- Require evaluators to explicitly verify the presence or absence of each described or claimed feature, step, or detail in the source material (e.g., image, code output, data), and to document any mismatches.\\
- Instruct evaluators to systematically check for calculation or logical errors in each answer, not just the reasoning structure, and to penalize answers with such errors even if the explanation appears logical.\\
- Encourage evaluators to reward answers that acknowledge ambiguity or uncertainty when the evidence is inconclusive, rather than making unsupported definitive claims.\\
- Require evaluators to compare all differences between answers, including minor or subtle distinctions, and to justify whether these differences materially affect the evaluation outcome.\\
- Instruct evaluators to reference authoritative sources or documentation (when applicable) to resolve disputes about standards, definitions, or expected outputs.\\
- Prompt evaluators to explicitly state any assumptions made during the evaluation and to consider whether those assumptions are justified by the evidence or context.\\
- Encourage evaluators to identify and explain any gaps in reasoning, missing steps, or lack of explicit justification in the answers, and to penalize accordingly.\\
- Require evaluators to ensure that their final justification is comprehensive, addressing every evaluation criterion and step, and not omitting any relevant aspect of the comparison.\\
\\
Output exactly ONLY THE PROMPT.
\end{tcolorbox}
\caption{Example of a meta prompt before an update from Figure~\ref{fig:meta_update_example}.}
\label{fig:meta_update_example_fig7_1}
\end{figure*}

\begin{figure*}[t]
\centering
\begin{tcolorbox}[breakable, width=0.95\textwidth, colback=blue!5, colframe=blue!50, boxrule=0.5pt, arc=2mm, title=Example of a meta prompt after an update]
Generate an evaluation prompt for an evaluator that includes example-specific evaluation criteria and step-by-step evaluation procedures. The evaluator will use this prompt to accurately judge the correctness and reasoning quality of given responses, and to identify which answer is better.\\
\\
The evaluation prompt you generate MUST include:\\
- Evaluation criteria tailored to the specific example, covering all aspects required for a thorough and precise judgment.\\
- Step-by-step evaluation procedures that are specific to the example and guide the evaluator to a reliable and accurate decision.\\
\\
Additionally, the evaluation prompt MUST instruct the evaluator that their final judgment must be expressed only in one of the following two formats:\\
`[[A]]' or `[[B]]'.\\
\\
When generating the evaluation prompt, incorporate the following additional tips to further improve accuracy and reliability (do NOT repeat tips already provided in the current meta prompt or previous feedback):\\
\\
**Additional Tips:**\\
- Require evaluators to explicitly verify the accuracy of arrangement, sequence, or spatial relationships described in each answer, especially when such details are visually or contextually important.\\
- Instruct evaluators to penalize both overstatements (e.g., exaggerating color, quantity, or significance) and omissions (e.g., failing to mention a visible but relevant feature), and to weigh the impact of these errors on the overall evaluation.\\
- Encourage evaluators to note when both answers contain errors or inaccuracies, and to justify the relative impact of those errors on the final decision, rather than defaulting to the answer with fewer mistakes.\\
- Prompt evaluators to reward answers that avoid unnecessary or unsupported inferences, especially when the evidence is ambiguous or incomplete.\\
- Require evaluators to check for and penalize the inclusion of details that are not supported by the source material, even if they do not directly contradict it.\\
- Instruct evaluators to briefly justify why the chosen answer best matches the evidence, even in cases where the answer is a single word or letter.\\
- Encourage evaluators to cite or reference authoritative sources, standards, or guidelines when resolving disputes about terminology, definitions, or expected outputs, and to document the source used.\\
- Remind evaluators to consider the clarity and explicitness of reasoning, rewarding answers that make their logic and assumptions transparent and penalizing those that leave reasoning steps implicit or unclear.\\
- Require evaluators to ensure that their final justification is not only comprehensive but also balanced, addressing both the strengths and weaknesses of each answer in relation to every evaluation criterion.\\
\\
Output exactly ONLY THE PROMPT.
\end{tcolorbox}
\caption{Example of a meta prompt after an update from Figure~\ref{fig:meta_update_example}.}
\label{fig:meta_update_example_fig7_2}
\end{figure*}

\end{document}